\documentclass[10pt,journal,compsoc]{IEEEtran}

\ifCLASSOPTIONcompsoc
  \usepackage[nocompress]{cite}
\else
  \usepackage{cite}
\fi

\ifCLASSINFOpdf
\else
\fi

\usepackage{amsmath}
\usepackage{graphicx,subfigure}
\usepackage{mathtools}
\usepackage{multirow}
\usepackage{amssymb}
\usepackage[pagebackref=true,breaklinks=true,letterpaper=true,colorlinks,bookmarks=false]{hyperref}

\def\eg{\emph{e.g.}}

\def\ie{\emph{i.e.}}
\def\etal{\emph{et al.}}

\newcommand{\ve}[1]{\mathbf{#1}} 
\newcommand{\ma}[1]{\mathrm{#1}} 

\newcommand{\fp}[2]{\frac{\partial{#1}}{\partial{#2}}}

\begin{document}
\title{BReG-NeXt: Facial Affect Computing Using Adaptive Residual Networks With Bounded Gradient}

\author{Behzad~Hasani,
        Pooran~Singh~Negi,
        and~Mohammad~H.~Mahoor,~\IEEEmembership{Senior Member,~IEEE}
\IEEEcompsocitemizethanks{\IEEEcompsocthanksitem Authors are with the Department
	of Electrical and Computer Engineering, University of Denver, Denver,
	CO, 80210.\protect\\
E-mail: \{bhasani, pooran.negi, mmahoor\}@du.edu
}
}

\markboth{IEEE Transactions on Affective Computing}
{Hasani \MakeLowercase{\textit{et al.}}: BReG-NeXt: Facial affect computing using adaptive residual networks with bounded gradient}

\IEEEtitleabstractindextext{
\begin{abstract}
This paper introduces \emph{BReG-NeXt}, a residual-based network architecture using a function wtih bounded derivative instead of a simple shortcut path (\emph{a.k.a.} identity mapping) in the residual units for automatic recognition of facial expressions based on the categorical and dimensional models of affect. Compared to ResNet, our proposed adaptive complex mapping results in a shallower network with less numbers of training parameters and floating point operations per second (FLOPs). Adding trainable parameters to the bypass function further improves fitting and training the network and hence recognizing subtle facial expressions such as contempt with a higher accuracy. We conducted comprehensive experiments on the categorical and dimensional models of affect on the challenging in-the-wild databases of AffectNet, FER2013, and Affect-in-Wild. Our experimental results show that our adaptive complex mapping approach outperforms the original ResNet consisting of a simple identity mapping as well as other state-of-the-art methods for Facial Expression Recognition (FER). Various metrics are reported in both affect models to provide a comprehensive evaluation of our method. In the categorical model, BReG-NeXt-50 with only 3.1M training parameters and 15 MFLOPs, achieves 68.50\% and 71.53\% accuracy on AffectNet and FER2013 databases, respectively. In the dimensional model, BReG-NeXt achieves 0.2577 and 0.2882 RMSE value on AffectNet and Affect-in-Wild databases, respectively.
\end{abstract}

}

\maketitle

\IEEEdisplaynontitleabstractindextext
\IEEEpeerreviewmaketitle
\IEEEraisesectionheading{\section{Introduction}\label{sec:introduction}}
\IEEEPARstart{A}{ffective} computing seeks to develop algorithms, systems and possibly devices that are capable of recognizing, interpreting, and simulating human emotions through different channels such as the face, voice, and biological signals~\cite{tao2005affective}. Facial expressions are the most important non-verbal channels used by human beings to convey internal feelings and emotions. There have been numerous efforts to develop robust and reliable automated Facial Expression Recognition (FER) systems that can understand human emotions and interact with subjects accordingly. However, currently available systems are far from reaching a comprehensive understanding of the emotional and social capabilities necessary for rich and robust Human Machine Interaction (HMI). This is predominantly due to the fact that HMI systems interact with humans in an uncontrolled environment (\emph{a.k.a.} wild settings), where the scene lighting, camera view, image resolution, background, subjects' head pose, gender, and ethnicity can vary significantly~\cite{c2,c53}.

Three models of categorical, dimensional, and Facial Action Coding System (FACS) are proposed in the literature to quantify affective facial behaviors:

\begin{figure}[tb]
	\centering
	\subfigure[Original Residual Unit]{\label{fig:ResNet}\includegraphics[width=0.45\linewidth]{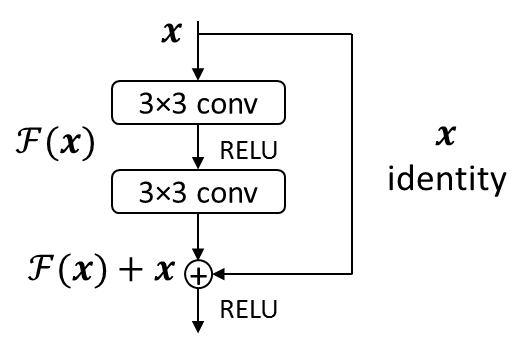}}
	\qquad
	\subfigure[BReG-NeXt Residual Unit]{\label{fig:BReG-NeXt}\includegraphics[width=0.45\linewidth]{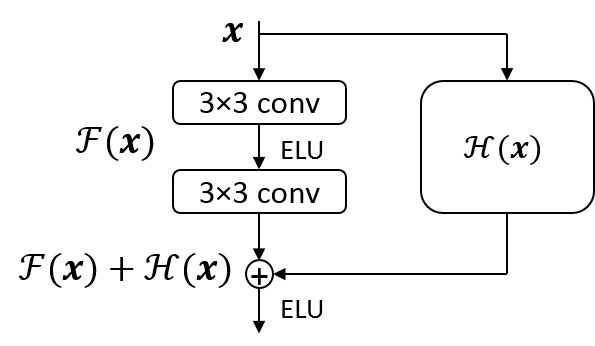}}
	\caption{Block diagram of a) Original Residual Unit b) BReG-NeXt Residual Unit (Equation\eqref{eq:arctanmapping} is used for $\mathcal{H}$)}\label{fig:blockdiagram}
\end{figure}

\begin{enumerate}
	\item \textbf{Categorical model}: Emotion is chosen from a list of affective-related categories that Ekman \emph{et al.}~\cite{c5} define as six basic emotions: anger, disgust, fear, happiness, sadness, and surprise.
	\item \textbf{Dimensional model}: A value is assigned to emotion over a continuous emotional
	scale, such as ``valence" and ``arousal", as defined by Russel~\cite{russell1980circumplex}, where valence shows how positive or negative an emotion is, and arousal indicates how much an event is intriguing or calming. This model can distinguish between subtle changes in exhibiting affect and encode these small differences in the intensity of each emotion on a continuous scale (\eg, \textit{valence} and \textit{arousal}).
	\item \textbf{FACS model}: All possible facial actions are described in terms of 33 Action Units (AUs)~\cite{ekman2002facial}. This model only describes facial movements and does not interpret the affective states directly.
\end{enumerate}

In traditional computer vision approaches, engineered features are used to describe visual patterns and build classifiers for visual object recognition. Alternately, Deep Neural Networks (DNNs) have the ability to learn and extract more discriminative features which yield in a better interpretation of the human face texture in visual data. Despite the superiority of DNNs over traditional methods, DNNs need a large amount of data for training the networks  properly. Thus, because of the small number of samples in the majority of the facial expression databases, training neural networks is significantly more difficult in this task~\cite{c53}.
 
In machine learning, one of the main goals is to optimally estimate a function or distribution with respect to a defined measure. Based on the connectionist principle~\cite{rumelhart1986parallel}, DNNs allow us to build very complex classes of functions. A tremendous number of network topologies have been proposed in recent years and they seem to play a crucial role in improving the underlying class of functions available to DNNs. In order to make the training of DNNs smoother and faster, current methods focus on improving neuron saturation or the efficiency of the gradient flow across various network's layers. Such approaches are more noticeable in the ReLU class of non-linear functions, and the use of identity mappings in Deep Residual Networks~\cite{he2016identity}. While having deeper architectures has shown to improve the result of classification~\cite{he2015deep}, one possibility is to design more complex neurons to extract more useful information at each layer of the network which results in shallower networks and fewer parameters to train but a more accurate extracted information and therefore a higher recognition rate.

In this work, we address the adaptive complex neuron concept by introducing \emph{BReG-NeXt} (following our previous work BReG-Net presented in~\cite{hasani2019bounded}).
In BReG-NeXt, instead of stacking up several identity-mapped residual units in hopes of increasing the number of parameters and eventually achieve a higher recognition rate (in architectures such as ResNet), we propose using a complex function to extract more information in each layer and therefore decrease the number of stacked layers and parameters (Figure~\ref{fig:blockdiagram}). In Section~\ref{sec:prposedmethod}, we will explain in detail that for our complex mapping formula we use:
\begin{equation}
\mathcal{H}(\ve{x}) = \frac{\tan^{-1}\left(\frac{\alpha\ve{x}}{\sqrt{\beta^2 + 1 }}\right)}{\alpha\sqrt{\beta^2 + 1 }}\nonumber
\end{equation}
where $\alpha$ and $\beta$ are trainable scalar parameters.

This mapping adds a small number of parameters to the original residual unit (two scalar parameters for each residual module) while at the same time extracts more useful features compared to the identity mapping proposed by He~\etal~\cite{he2015deep}. One important feature of this mapping is that its gradient is bounded and continues on $\ve{x} \in \mathbb{R}$ (when $\alpha \in \mathbb{R} - \{0\}$ and $\beta \in \mathbb{R}$). Therefore, it preserves all the properties of identity mapping and at the same time it prevents the exploding or vanishing gradient problem during backpropagation. The entire network can still be trained end-to-end by ADAM optimizer with backpropagation and can be easily implemented using common DNN libraries.

We conduct comprehensive experiments on three in-the-wild facial expression databases (AffectNet~\cite{mollahosseini2017affectnet}, Affect-in-the-wild~\cite{zafeiriou2017aff}, and FER2013~\cite{FER2013}) to evaluate our method. Our experiments show that BReG-NeXt architecture significantly reduces the number of parameters compared to the identity mapping residual networks while produces better prediction rates on both categorical and dimensional models of affect. For the categorical model, we achieve \textbf{68.50\%} and \textbf{71.53\%} recognition rate on  AffectNet and FER2013, respectively. For the dimensional model, we achieve \textbf{0.2577} and \textbf{0.2882} RMSE on AffectNet and Affect-in-Wild, respectively. These results outperform many state-of-the-art  methods evaluated on these databases so far.

The remainder of the paper is organized as follows: Section~\ref{sec:relatedwork} provides an overview of the related work in this field. Section~\ref{sec:prposedmethod} explains the BReG-NeXt architecture and the functions utilized to build the network. Section~\ref{sec:expresults} presents experimental results and their analysis and finally, Section~\ref{sec:conclusion} concludes the paper with some discussions and recommendations for future research.

\section{Related Work}
\label{sec:relatedwork}
In this section, we overview related work in Facial Expression Recognition on the categorical and dimensional models of affect. We also mention recent findings in regards to incorporating more complex nodes in DNNs.
\subsection{Facial Expression Recognition on Categorical Model of Affect}
Traditional approaches for automated affective computing use various engineered features such as Local Binary Patterns (LBP)~\cite{c2}, Histogram of Oriented Gradients (HOG)~\cite{c14},  Histogram of Optical Flow (HOF)~\cite{c15}, and facial landmark points~\cite{c12}. These engineered features often times lack the required generalizability power that makes the method robust to high variation in important factors such as lighting, views, resolution, subjects' ethnicity, etc.
Deep learning has become a hot research topic and has achieved state-of-the-art performance for a variety of applications~\cite{deng2014deep} as well as facial affect estimation. In this section, we briefly mention some of the deep learning-based methods used for FER.

\begin{figure*}[tbh]
	\centering
	\includegraphics[width=\linewidth]{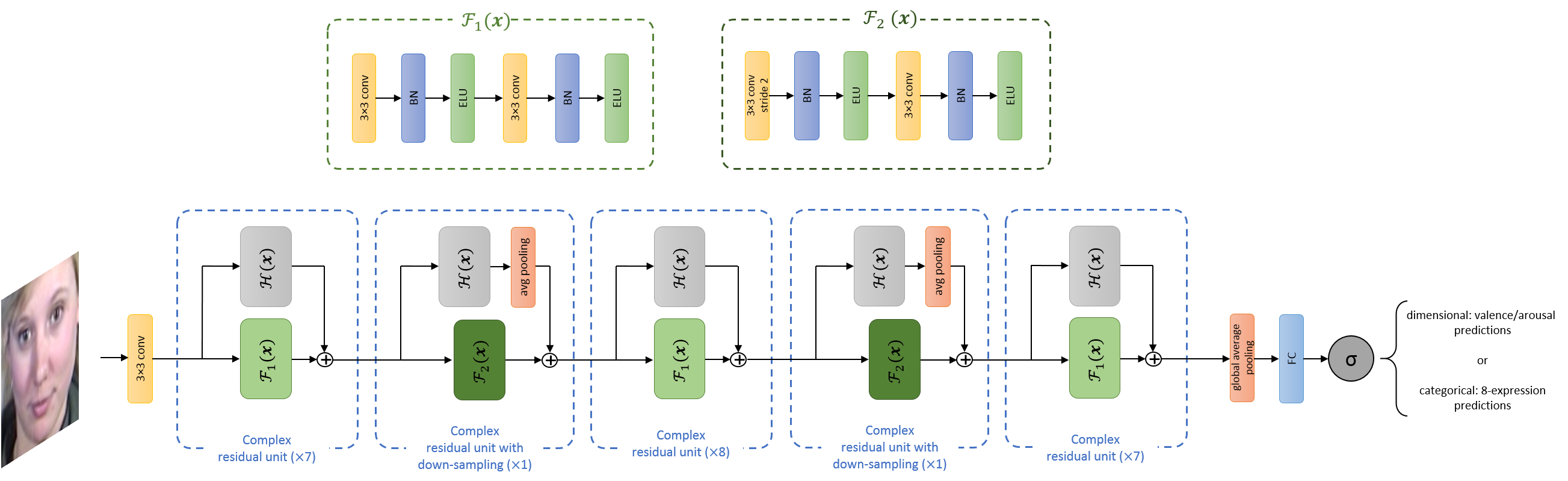}
	\caption{Network configuration of BReG-NeXt-50. Down-sampling on the bypass route is implemented by average pooling. $\mathcal{H}$ is implemented with Equation~\eqref{eq:arctanmapping}}\label{fig:overview}
\end{figure*}

Human emotion can be recognized using audio and visual information that express different non-verbal cues such as language, gesture, and facial expression. These modalities are either used individually or in combination. Although categorizing expressions based on visual data can achieve promising results, incorporating other models can provide extra information and further enhance the recognition rate. For instance, in the EmotiW and Audio Video Emotion Challenges (AVEC)~\cite{valstar2016avec,ringeval2017avec}, the audio model was considered to be the second most important element. Various fusion techniques for multi-modal affect recognition were proposed in these challenges. Li \etal~\cite{li2017multimodal} proposed a deep fusion CNN (DF-CNN) to explore multi-modal 2D+3D FER. Specifically, six types of 2D facial attribute maps (\ie, geometry, texture, curvature, normal components $x$, $y$, and $z$) were first extracted from the textured 3D face scans and then were jointly fed into the feature extraction and feature fusion subnets to learn the optimal combination weights of 2D and 3D facial representations.
Also, Vielzeuf \etal~\cite{vielzeuf2017temporal} proposed a multi-modal approach for video emotion classification by combining VGG and C3D
models as image descriptors.

 In other DNN-based methods, Mollahosseini \emph{et al.}~\cite{c4,c53} have used the Inception layer for the task of facial expression recognition and achieved significant results. Moreover, Inception layer is combined with a residual unit introduced by He \emph{et al.}~\cite{he2015deep}. They showed that the resulting architecture accelerates the training of Inception networks significantly~\cite{c23}. ResNet-based methods have been extensively investigated in the literature~\cite{hu2017learning,hasani2017facial,hasani2017facial_dimensional,hasani2019bounded} and have shown significant results in FER. Hasani \emph{et al.} proposed a modification of ResNets for the task of facial expression recognition~\cite{hasani2017facial} and valence/arousal prediction of emotions~\cite{hasani2017facial_dimensional}. Many of these methods use very deep architectures that required training millions of parameters as well as a considerable amount of memory and computation power to train them. Therefore, the main question here is whether having a more complex building block of neural networks results in a shallower and more efficient network or not? In this work, we address this question and investigate the impact of this concept.

\subsection{Dimensional Model of Affect}
Traditional methods for visual prediction of the dimensional model of affect have been a topic of study for years. Gunes~\etal~\cite{gunes2010dimensional} focus on the dimensional prediction of emotions from spontaneous conversational head gestures by mapping the amount and direction of head motion
and occurrences of head nods and shakes into arousal, expectation,
intensity, power and valence level of the observed subject using Support Vector Regressions
(SVRs). Kipp~\etal~\cite{kipp2009gesture} investigated (without performing automatic prediction) how basic gestural form features (\eg,
preference for using the left/right hand, hand shape, palm orientation,
etc.) are related to the single Pleasure-Arousal-Dominance (PAD)~\cite{mehrabian1996pleasure} dimensions of emotion. Nicolaou~\etal~\cite{nicolaou2012output} focus on the dimensional and continuous
prediction of emotions from naturalistic facial expressions within
an Output-Associative Relevance Vector Machine (RVM) regression
framework by learning non-linear input and output dependencies
inherent in the affective data. In~\cite{nicolaou2010automatic} a novel technique to automatically segment emotional clips from
long audiovisual interactions is proposed. Also in~\cite{arifin2008affective} extracting
emotional segments from video based on the PAD model (assuming independency between the dimensions) is introduced.

As mentioned before, fewer studies have been conducted on the dimensional model of affect using DNNs as there are not many datasets with a large number of images available in this area.  Nicolaou \emph{et al.}~\cite{nicolaou2011continuous} trained bidirectional Long Short Term Memory (LSTM) architecture on multiple engineered features extracted from audio, facial geometry, and shoulders. They achieved Root Mean Square Error (RMSE) of 0.15 and Correlation Coefficient (CC) of 0.79 for valence as well as RMSE of 0.21 and CC of  0.64 for arousal. 

 He~\emph{et al.}~\cite{he2015multimodal} won the AVEC 2015 challenge by training multiple stacks of bidirectional LSTMs (DBLSTM-RNN) on engineered features extracted from audio (LLDs features), video (LPQ-TOP features), 52 ECG features, and 22 EDA features. They achieved RMSE of 0.104 and CC of 0.616 for valence as well as RMSE of 0.121 and CC of 0.753  for  arousal. Koelstra \emph{et al.}~\cite{koelstra2012deap} trained Gaussian naive Bayes classifiers  on EEG, physiological signals, and multimedia features by binary classification of low/high categories for arousal, valence, and liking on their proposed database DEAP. They achieved F1-score of 0.39, 0.37, and 0.40 on arousal, valence, and liking categories respectively. Authors in~\cite{hewitt2018cnn} propose three CNN-based facial affect prediction method for mobile devices. In~\cite{wang2018two} a two-level attention
 with two-stage multi-task learning framework is proposed for facial emotion
 estimation on static images using Bi-directional Recurrent Neural Networks (Bi-RNNs). In~\cite{langholz2019oculum} a CNN-based method is proposed for predicting valence and arousal in images by focusing on the ocular region.  Same as methods in the categorical model of affect, these methods are very deep networks with very high numbers of parameters to train. 

\subsection{More Complex Nodes in Residual Networks}
Having more complex units in residual networks has not been thoroughly investigated in the literature. This might be partially due to the fact that He~\etal~in~\cite{he2016identity} argues that by having the mapping $\mathcal{H}(\ve{x}_l) = \lambda_{l}\ve{x}_l$  instead of the shortcut bypass (see Figure~\ref{fig:blockdiagram}), then for large values for $\lambda_{l}$, the gradient in backpropagation will be exponentially large and for small values of $\lambda_{l}$, it would be exponentially small and therefore the gradient vanishes. We will address this concern in the next section. 

Our initial experiments for having a complex mapping in residual units in~\cite{hasani2019bounded} (BReG-Net) showed that by having $\mathcal{H}(\ve{x}_l) = \tan^{-1}(\ve{x}_l)$, that has a bounded and continues gradient on $\ve{x}_l \in \mathbb{R}$, not only do we prevent from facing vanishing/exploding gradient problem, but we have much less number of parameters to learn and also the network converges considerably faster than using the original identity mapping. Based on this work, we investigated more general forms of functions in the residual units as well as making the mapping adaptive to the input data by introducing trainable parameters for each residual unit as explained in Section~\ref{sec:prposedmethod}.

\begin{figure*}[tbh]
	\centering
	\subfigure[$\alpha = 1$, $\beta = 1$]{\label{a=1,b=1}\includegraphics[width=0.24\linewidth]{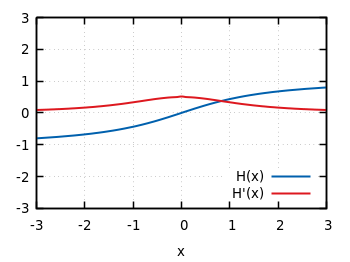}}
	\subfigure[$\alpha = 1$, $\beta = 0$]{\label{a=1,b=0}\includegraphics[width=0.24\linewidth]{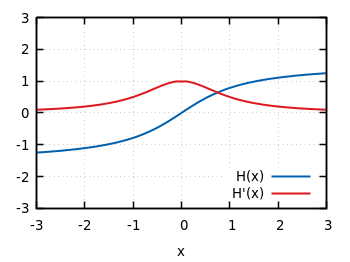}}
	\subfigure[$\alpha \to 0$, $\beta = 1$]{\label{a=0,b=1}\includegraphics[width=0.24\linewidth]{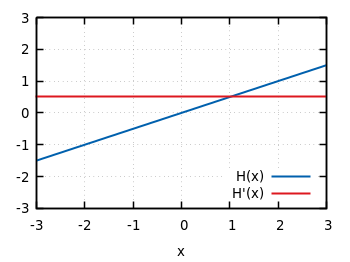}}
	\subfigure[$\alpha \to 0$, $\beta = 0$]{\label{a=0,b=0}\includegraphics[width=0.24\linewidth]{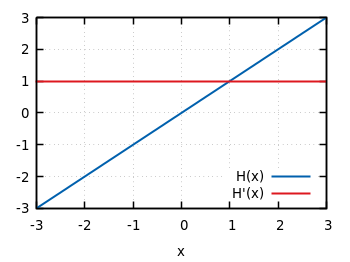}}
	\caption{Plots of proposed complex mapping function $\mathcal{H}$ (Equation~\eqref{eq:arctanmapping}) and its derivative ($\mathcal{H}'$) for different values of $\alpha$ and $\beta$ (best viewed in color)}\label{fig:functionPlots}
\end{figure*}

\section{BReG-NeXt}\label{sec:prposedmethod}

\subsection{Residual Networks}

Deep Residual Networks (ResNets)~\cite{he2015deep} are shaped by stacking several residual building block units. Each of these units can be mathematically shown as:
\begin{equation}
\begin{gathered}
\ve{y}_{l} = \mathcal{H}(\ve{x}_{l}) + \mathcal{F}(\ve{x}_{l}, \mathcal{W}_l)\\
\ve{x}_{l+1} = f(\ve{y}_{l})
\end{gathered}
\label{eq:resnetformula}
\end{equation}
where $\ve{x}_{l}$ and $\ve{x}_{l+1}$ are input and output of the $l$-th unit, $\mathcal{F}$ is a residual function, and $\mathcal{W}_l=\{\ma{W}_{l,k} | _{1\leq k \leq K}\}$ is a set of weights (and biases) associated with the $l$-th Residual Unit in which $K$ is the number of layers in a Residual Unit. In~\cite{he2015deep}, $\mathcal{H}(\ve{x}_{l}) = \ve{x}_{l}$ is an identity mapping and $f$ is a ReLU activation function. Very deep ResNets have shown state-of-the-art recognition rates for several challenging classification and detection tasks on ImageNet~\cite{russakovsky2015imagenet} and MS COCO~\cite{lin2014microsoft} competitions. The main idea behind ResNets revolves around learning the additive residual function $\mathcal{F}$ with respect to $\mathcal{H}(\ve{x}_{l})$, where  $\mathcal{H}(\ve{x}_{l}) = \ve{x}_{l}$.

Thus, \textit{Residual Units} are formulated as Equation~\eqref{eq:resnetformula}. In this equation,
if $\mathcal{H}$ is an identity mapping, and $f$ is an identity function, then we will have:
\begin{equation}
\ve{x}_{l+1} = \ve{x}_{l} + \mathcal{F}(\ve{x}_{l}, \mathcal{W}_{l}). \label{eq:additive0}
\end{equation}

For backpropagation, with $\mathcal{E}$ as loss function we will have:
\begin{equation}
\begin{aligned}
\fp{\mathcal{E}}{\ve{x}_{l}}&=\fp{\mathcal{E}}{\ve{x}_{L}}\fp{\ve{x}_{L}}{\ve{x}_{L-1}}\dots\fp{\ve{x}_{l+1}}{\ve{x}_{l}}\\
&=\fp{\mathcal{E}}{\ve{x}_{L}}\prod_{i=l}^{L-1}\left(\fp{\mathcal{F}(\ve{x}_i,\mathcal{W}_i)}{\ve{x}_{i}} + 1\right)\label{eq:grad}
\end{aligned}
\end{equation}
for any deeper unit $L$ and any shallower unit $l$.

\subsection{Complex Mapping}
\label{sec:identityskip}
If we replace the identity mapping with a more complex function of $\mathcal{H}_l(\ve{x}_l)$, assuming $f$ remains an identity function, then Equation~\eqref{eq:resnetformula} will be:
\begin{equation}
\ve{x}_{l+1} = \mathcal{H}_l(\ve{x}_{l}) + \mathcal{F}(\ve{x}_{l}, \mathcal{W}_l)
\end{equation}
By recursively applying this formulation and then calculating the backpropagation, similar to Equation~\eqref{eq:grad} we will have:

\begin{equation}
\begin{aligned}
\fp{\mathcal{E}}{\ve{x}_{l}}&=\fp{\mathcal{E}}{\ve{x}_{L}}\fp{\ve{x}_{L}}{\ve{x}_{L-1}}\dots\fp{\ve{x}_{l+1}}{\ve{x}_{l}}\\
&=\fp{\mathcal{E}}{\ve{x}_{L}}\prod_{i=l}^{L-1}\left(\fp{\mathcal{F}(\ve{x}_i,\mathcal{W}_i)}{\ve{x}_{i}} + \fp{\mathcal{H}_i(\ve{x}_i)}{\ve{x}_{i}}\right)\label{eq:grad1}
\end{aligned}
\end{equation}

In~\cite{he2016identity} a simple modification of  $\mathcal{H}(\ve{x}_{l}) = \lambda_l\ve{x}_{l}$ is investigated (where $\lambda_l$ is a modulating scalar). By putting this function in Equation~\eqref{eq:grad1}, we will have:

\begin{equation}
\begin{aligned}
\fp{\mathcal{E}}{\ve{x}_{l}}=
\fp{\mathcal{E}}{\ve{x}_{L}}\prod_{i=l}^{L-1}\left(\fp{\mathcal{F}(\ve{x}_i,\mathcal{W}_i)}{\ve{x}_{i}} + \lambda_i\right)\label{eq:grad2}
\end{aligned}
\end{equation}

In this case, for a very deep network (large $L$), if $\lambda_{i}>1$ for all $i$, this factor can be considerably large. Also, if $\lambda_{i}<1$ for all $i$, this factor can be exponentially small and therefore it vanishes, which blocks the backpropagated signal from the shortcut and forces it to flow through the weight layers. Therefore, choosing a suitable replacement for $\mathcal{H}(\ve{x})$ is very critical for  the network's convergence. In an ideal case, all of the properties of the identity mapping function are needed to be preserved. One of the main properties of identity mapping is that it is continuous on $\mathbb{R}$ and it is also bounded (always equal to 1). From Equation~\eqref{eq:grad2} we realize that the value of the derivative of the mapping function needs to be bounded and ideally less than or equal to 1. Also, this value should not be very small because similar to the case $\mathcal{H}(\ve{x}_{l}) = \lambda_l\ve{x}_{l}$, in very deep networks the gradient would vanish along the bypass path.

Based on the argument above, we investigated several functions (adaptive and non-adaptive) with bounded derivatives. Few of these functions are as follows:

\begin{equation}
\begin{split}
\mathcal{H}_1(\ve{x})&=\tan^{-1}(\ve{x}),~\fp{\mathcal{H}_1}{\ve{x}}=\frac{1}{1+\ve{x}^2}\\
\mathcal{H}_2(\ve{x})&= \ve{x}\tan^{-1}(\ve{x})-\frac{1}{2}\log(\ve{x}^2+1),~\fp{\mathcal{H}_2}{\ve{x}}= \tan^{-1}(\ve{x}) \\
\mathcal{H}_3(\ve{x})&= - \frac{\log(e^{\ve{x}} +\alpha^2)}{\alpha^2},~\fp{\mathcal{H}_3}{\ve{x}}=\frac{1}{e^{\ve{x}} + \alpha^2}
\end{split}
\label{eq:modification_funnctions}
\end{equation} 
However, our experiments showed the best results with the following mapping:
\begin{equation}
\mathcal{H}(\ve{x}_l,\alpha_l,\beta_l) = \frac{\tan^{-1}\left(\frac{\alpha_l\ve{x}_l}{\sqrt{\beta_l^2 + 1}}\right)}{\alpha_l\sqrt{\beta_l^2 + 1}}\label{eq:arctanmapping}
\end{equation}
where $\alpha_l$ and $\beta_l$ are trainable scalars for the $l$-th layer of the residual unit. This mapping is based on the result of our initial findings in~\cite{hasani2019bounded} for BReG-Net. In the following we explain different aspects of this mapping.

First, Equation~\eqref{eq:arctanmapping} is continuous and differentiable on $\ve{x}_l\in \mathbb{R}$ (for $\alpha_l \in \mathbb{R} - \{0\}$ and $\beta_l \in \mathbb{R}$), which means that it preserves those properties of identity mapping. Second, its partial derivative over $\ve{x}_l$ is bounded:
\begin{equation}
\fp{\mathcal{H}(\ve{x}_l,\alpha_l,\beta_l)}{\ve{x}_l} = \frac{1}{\alpha_l^2\ve{x}_l^2 + \beta_l^2 + 1}\label{eq:arctangrad}
\end{equation}
therefore, $\forall \ve{x}_l, \alpha_l, \beta_l \in \mathbb{R}: 0 < \fp{\mathcal{H}(\ve{x}_l,\alpha_l,\beta_l)}{\ve{x}_l} \leq 1, $ which means that it preserves that property of identity mapping as well. Third, to prevent the exploding gradient problem (as shown in Equation~\eqref{eq:grad2}) we prefer a function that its derivative is not above 1. $\mathcal{H}$ satisfies this condition as well. Fourth, for reasonable values for $\alpha_l$ and $\beta_l$ (which is the case for almost all of the training scenarios), $\fp{\mathcal{H}}{\ve{x}_l}$ is far from becoming zero (especially when batch normalization is applied and the data is zero-centered). Therefore, it is very unlikely for the residual unit to face the vanishing gradient problem even for very deep networks. Figure~\ref{fig:functionPlots} shows the plots for $\mathcal{H}$ and its derivative for different values of $\alpha$ and $\beta$. It can be seen that it is very unlikely for $\mathcal{H}^{'}$ to have near-zero value as input $\ve{x}_l$ is mostly around zero after batch normalization.   By putting all of the equations together our proposed complex mapping and the backpropagation will be as follows:
\begin{equation}
\begin{gathered}
\ve{x}_{l+1} =  \frac{\tan^{-1}\left(\frac{\alpha_l\ve{x}_l}{\sqrt{\beta_l^2 + 1}}\right)}{\alpha_l\sqrt{\beta_l^2 + 1}} + \mathcal{F}(\ve{x}_{l}, \mathcal{W}_l) \\
\fp{\mathcal{E}}{\ve{x}_{l}}=\fp{\mathcal{E}}{\ve{x}_{L}}\prod_{i=l}^{L-1}\left(\fp{\mathcal{F}(\ve{x}_i,\mathcal{W}_i)}{\ve{x}_{i}} + \frac{1}{\alpha_i^2\ve{x}_i^2 + \beta_i^2 + 1}\right)
\label{eq:grad3}
\end{gathered}
\end{equation}

Equation~\eqref{eq:grad3} shows that our proposed mapping flows the gradient smoothly in the backpropagation and addresses the concerns in~\cite{he2016identity} for complex mappings. By replacing the original identity mapping with the proposed $\mathcal{H}$ we will have more complex nodes in our residual neural network which results in having shallower networks and therefore fewer number of parameters to train as we show by our experiments.

\begin{table*}[htbp]
	\centering
	\caption{Architecture of studied networks in this work. The provided values for convolution layers are size of the convolution filters followed by number of their output channel. The provided information for fully connected layer is their output size followed by activation function. }
	\label{tab:arch}
	\resizebox{\textwidth}{!}{
	\begin{tabular}{cc|c|c|c|c|c}
		\cline{2-7}
		& \textbf{ResNet-32} & \textbf{ResNet-50}& \textbf{BReG-Net-32} & \textbf{BReG-Net-50} & \textbf{BReG-NeXt-32} & \textbf{BReG-NeXt-50}\\ \hline \hline
		\multicolumn{1}{c|}{conv1}          & \multicolumn{2}{c|}{3$\times$3, 64, stride 2} &  \multicolumn{4}{c}{3$\times$3, 32}               \\ \hline
		\multicolumn{1}{c|}{conv2}          & $\left[ \begin{array}{c} 3\times3,64  \\3\times3,64\end{array}\right] \times 3$ & $\left[ \begin{array}{c} 3\times3,64  \\3\times3,64\end{array}\right] \times 8$ & $\left[ \begin{array}{c} 3\times3,32  \\3\times3,32\end{array}\right] \times5$  & $\left[ \begin{array}{c} 3\times3,32  \\3\times3,32\end{array}\right] \times8$  & $\left[ \begin{array}{c} 3\times3,32  \\3\times3,32\end{array}\right] \times4$ & $\left[ \begin{array}{c} 3\times3,32  \\3\times3,32\end{array}\right] \times7$ \\ \hline
		\multicolumn{1}{c|}{conv3}          & $\left[ \begin{array}{c} 3\times3,128  \\3\times3,128\end{array}\right] \times 3$ & $\left[ \begin{array}{c} 3\times3,128  \\3\times3,128\end{array}\right] \times 1$ & $\left[ \begin{array}{c} 3\times3,64  \\3\times3,64\end{array}\right] \times 1$ & $\left[ \begin{array}{c} 3\times3,64  \\3\times3,64\end{array}\right] \times 1$ &$\left[ \begin{array}{c} 3\times3,64  \\3\times3,64\end{array}\right] \times 1$  & $\left[ \begin{array}{c} 3\times3,64  \\3\times3,64\end{array}\right] \times 1$ \\ \hline
		\multicolumn{1}{c|}{conv4}          & $\left[ \begin{array}{c} 3\times3,256  \\3\times3,256\end{array}\right] \times 5$& $\left[ \begin{array}{c} 3\times3,128  \\3\times3,128\end{array}\right] \times 7$ & $\left[ \begin{array}{c} 3\times3,64  \\3\times3,64\end{array}\right] \times4$ & $\left[ \begin{array}{c} 3\times3,64  \\3\times3,64\end{array}\right] \times7$ &$\left[ \begin{array}{c} 3\times3,64  \\3\times3,64\end{array}\right] \times5$ & $\left[ \begin{array}{c} 3\times3,64  \\3\times3,64\end{array}\right] \times8$ \\ \hline
		\multicolumn{1}{c|}{conv5}          & $\left[ \begin{array}{c} 3\times3,512  \\3\times3,512\end{array}\right] \times 3$& $\left[ \begin{array}{c} 3\times3,256  \\3\times3,256\end{array}\right] \times 1$ & $\left[ \begin{array}{c} 3\times3,128  \\3\times3,128\end{array}\right] \times1$ & $\left[ \begin{array}{c} 3\times3,128  \\3\times3,128\end{array}\right] \times1$ &$\left[ \begin{array}{c} 3\times3,128  \\3\times3,128\end{array}\right] \times1$& $\left[ \begin{array}{c} 3\times3,128  \\3\times3,128\end{array}\right] \times1$\\ \hline
		\multicolumn{1}{c|}{conv6}          & -&$\left[ \begin{array}{c} 3\times3,256  \\3\times3,256\end{array}\right] \times 7$& $\left[ \begin{array}{c} 3\times3,128  \\3\times3,128\end{array}\right] \times 4$ & $\left[ \begin{array}{c} 3\times3,128  \\3\times3,128\end{array}\right] \times 7$ & $\left[ \begin{array}{c} 3\times3,128  \\3\times3,128\end{array}\right] \times 4$ & $\left[ \begin{array}{c} 3\times3,128  \\3\times3,128\end{array}\right] \times 7$\\ \hline
		\multicolumn{1}{c|}{\begin{tabular}[c]{@{}c@{}}global \\ avg pooling\\  + \\ fully connected\end{tabular}}   & \multicolumn{6}{c}{8, softmax (categorical) / 2, linear (dimensional)}\\ \hline \hline
		\multicolumn{1}{c|}{\begin{tabular}[c]{@{}c@{}}number of \\ prameters\end{tabular}} & 19.6M                 & 25M                & 1.9M    & 3.1M & 1.9M   & 3.1M             \\ \hline
		\multicolumn{1}{c|}{FLOPs}          & $9.8\times10^7 $& $12.5\times10^7$  & $0.93\times10^7$ & $1.51\times10^7$ & $0.95\times10^7 $ & $1.53\times10^7$\\ \hline
	\end{tabular}
}
\end{table*}

\subsection{Adaptive Mapping}
The reason behind defining $\alpha_l$ and $\beta_l$ in Equation~\eqref{eq:arctanmapping} is to make each residual unit fit its own input and adjust the complex mapping accordingly. This is vital for facial affect estimation task where subtle changes in the input data are needed to be detected and recognized. Furthermore, in~\cite{he2016identity} it has been shown that having training parameters in the bypass (\eg, exclusive gating, shortcut-only gating, etc.) reduces the error rate comparing to only scaling the bypass without involving any training parameters. We discovered the same phenomena in our experiments where by assigning $\forall i \in \mathbb{N} : \alpha_i = 1,~\beta_i = 0$ (\ie,~$\mathcal{H}(\ve{x}) = \tan^{-1}(\ve{x})$) error rates increased as shown by our experiments.

\subsection{Network Architecture}
As mentioned before, we have tested various mappings and observed consistent results for our proposed complex mapping. In order to compare the effectiveness of our method, we investigate six  networks (three shallow and three deep architectures): 1)~ResNet-32  2)~ResNet-50 3)~BReG-Net-32 (BReG-Net is a special case of BReG-NeXt with $\alpha = 1$  and $\beta = 0$ in Equation~\ref{eq:arctanmapping}) 4)~BReG-Net-50 5) BReG-NeXt-32 which is comparable with ResNet-32 and BReG-Net-32 in terms of number of layers 6)~BReG-NeXt-50 which is our final proposed architecture that achieved best results on both categorical and dimensional models of affect and is comparable with ResNet-50 and BReG-Net-50.

\vspace{10pt}
\noindent\textbf{ResNet-32:} Our first baseline is ResNet-32 proposed by He \emph{et al.} in~\cite{he2015deep} where identity mapping is used for the bypass over the $3\times3$ convolutions. The detailed structure of this baseline is provided in Table~\ref{tab:arch}. Our implementation of this network is slightly different from the one mentioned in~\cite{he2015deep} as we intended to make this network similar to BReG-NeXt in terms of the arrangement of residual units to have a fair comparison between the two architectures.

\vspace{10pt}
\noindent\textbf{ResNet-50:} Our second baseline is ResNet-50 which is a deeper version of ResNet-32. Similar to the previous network, for a fair comparison, our implementation of this network is slightly different from~\cite{he2015deep}.

\vspace{10pt}
\noindent\textbf{BReG-Net-32:} The next baseline is BReG-Net-30~\cite{hasani2019bounded}. BReG-Net was our first attempt in developing residual units with complex mapping. BReG-Net simply uses $\mathcal{H}(\ve{x}) = \tan^{-1}(\ve{x})$ for its mapping function with no additional training parameters. In other words, BReG-Net uses the complex mapping of Equation~\ref{eq:arctanmapping} while $\alpha = 1$ and $\beta = 0$ are always fixed for all blocks. The number of training parameters for  this architecture is 1.9M which is significant reduction compared to the previously mentioned ResNets (Table~\ref{tab:arch}). Therefore, it is a suitable point of reference for our proposed complex mapping for BReG-NeXt. Similar to the previous networks, for a fair comparison, we reduced the layers of BReG-Net originally proposed in~\cite{hasani2019bounded} as the original architecture contains more number of layers compared to its BReG-NeXt counterpart.

\vspace{10pt}
\noindent\textbf{BReG-Net-50:} Our next baseline is a deeper version of BReG-Net to compare the deep versions of the architectures. Similar to the previous networks, for a fair comparison, we matched the number of layers of BReG-Net to its BReG-NeXt counterpart.

\vspace{10pt}
\noindent\textbf{BReG-NeXt-32:} Our shallow version of BReG-NeXt has 32 layers and is comparable to ResNet-32 and BReG-Net-32 in terms of depth. In terms of the number of training parameters, however, BReG-NeXt-32 is significantly lighter than ResNet-32 with only 1.9M parameters (Table~\ref{tab:arch}). In this architecture, down-sampling is applied after the complex mapping $\mathcal{H}$ at the  same time that the number of feature map channels increases (conv3 and conv5 in Table~\ref{tab:arch}).

\vspace{10pt}
\noindent\textbf{BReG-NeXt-50:} Our deeper version of BReG-NeXt has 50 layers (Figure~\ref{fig:overview}). In the experiments section, we show that this network achieves the best results in both categorical and dimensional models of affect by having only 3.1M trainable parameters.

Table~\ref{tab:arch} provides a general overview of the six networks that we study in this paper. ResNet-32, BReG-Net-32, and BReG-NeXt-32 are similar networks in terms number of convolution layers and they are a shallower version of their architectures while ResNet-50, BReG-Net-50, and BReG-NeXt-50 are a deeper version of them. ResNets showed significant results in~\cite{he2015deep} therefore they are suitable benchmarks for our method. Considering the parameter/error-rate trade-off, we propose BReG-NeXt-50 as our final network since deeper networks did not show significant improvement as it is shown in the experiments section.

\subsection{Implementation}
\label{sec:impl}

We implemented our method using a combination of TensorFlow~\cite{AABB01}, TfLearn~\cite{tflearn2016}, and Keras~\cite{chollet2015keras} libraries. For all experiments on all databases, we crop the faces and resize them to 64$\times$64$\times$3 pixels. For augmentation, random horizontal flip is used followed by random changes in hue, saturation, brightness, contrast, and zooming. Augmentation is applied in 25\% of the time. Zero-centering for each color channel is also utilized as the gradient flows more smoothly around zero.   
Similar to ResNet, we use batch normalization~\cite{c20} after each convolution and before each activation. We use ``ELU" activation function~\cite{clevert2015fast} as it contributes to flowing the gradient more smoothly for negative values in the backpropagation. All networks are trained from scratch. Focal loss is used for the categorical model experiments and Mean Squared Error loss is used for the dimensional model experiments. ADAM optimizer with the batch size of 128 is used for all experiments. The learning rate starts from 0.0001 and is multiplied by 0.8 after every 10 epochs. Similar to ResNet we do not use dropout~\cite{he2015deep,c20}, however, we use L2 regularizers on the convolution layers. Our code and trained network parameters will be made publicly available.

\begin{figure*}[th]
	\centering
	\subfigure[AffectNet]{\label{fig:AffectnetExamples}\includegraphics[width=0.25\linewidth]{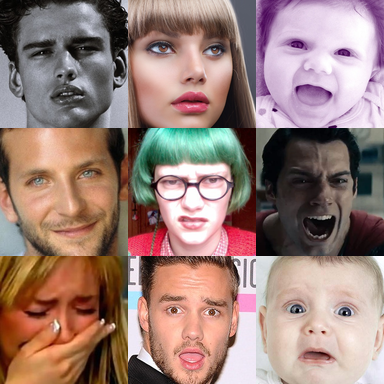}}
	\qquad
	\subfigure[FER2013]{\label{fig:FER2013Examples}\includegraphics[width=0.25\linewidth]{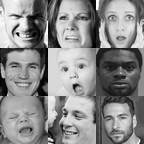}}
	\qquad
	\subfigure[Affect-in-Wild]{\label{fig:aff-in-wildExamples}\includegraphics[width=0.25\linewidth]{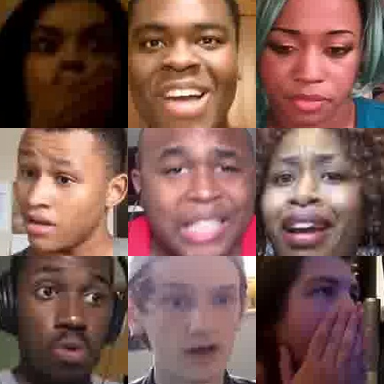}}
	\caption{Some example images from AffectNet (a), FER2013 (b), and Affect-in-Wild (c) databases used in this study. AffectNet contains labels for both categorical and dimensional models of affect while FER only contains labels for categorical model and Affect-in-Wild only contains labels for dimensional model.}\label{fig:databaseExamples}
\end{figure*}

\section{Experiments}\label{sec:expresults}

In this section, we briefly review the three face databases used for evaluating our proposed method. We then provide details of our experiments and their results using these databases evaluated on different metrics on both categorical and dimensional models of affect.

\subsection{Face Databases}
As noted earlier, many of the traditional facial expression databases are assembled in a controlled environment while for developing a practical method, these databases do not yield satisfying results. Therefore, we chose databases that are captured in the wild setting which contain a variety of backgrounds, lighting, pose, subject ethnicity, etc. These databases are AffectNet~\cite{mollahosseini2017affectnet}, Affect-in-Wild~\cite{zafeiriou2017aff}, and FER2013~\cite{FER2013} of which AffectNet contains labels of both categorical and dimensional models. Affect-in-Wild contains only labels of the dimensional model, and FER2013 contains only labels of the categorical model. In the following, we briefly review the contents
of these databases.

\vspace{10pt}
\noindent\textbf{AffectNet} contains more than one million facial images collected from the Internet by querying three major search engines using 1250 emotion related keywords in six different languages~\cite{mollahosseini2017affectnet}. About half of the retrieved images (around 440,000 images) are manually annotated for the presence of seven discrete facial expressions (categorical model) and the intensity of valence and arousal (dimensional model). AffectNet is the largest database of facial expressions, valence, and arousal in the wild enabling research in automated facial expression recognition in two different emotion models. This database is very challenging as it contains images of people from different races and ethnicities as well as high variety in the background, lighting, pose, point of view, etc. Figure~\ref{fig:AffectnetExamples} shows some example images from AffectNet database.

\vspace{10pt}
\noindent\noindent\textbf{FER2013} was  introduced  in  the  ICML  2013 Challenges in Representation Learning~\cite{FER2013}. The database was created using the Google image search API and faces have  been  automatically  registered.   Faces  are  labeled with any of the six basic expressions, along with neutral. The resulting database contains 35,887 images in wild settings. Few examples of this database are provided in Figure~\ref{fig:FER2013Examples}.

\vspace{10pt}
\noindent\textbf{Affect-in-Wild} was introduced in CVPR 2017 workshop challenge~\cite{zafeiriou2017aff}. 
This database contains 300 videos of different subjects watching videos of various TV shows and movies. The videos contain subjects from different genders and ethnicities with high variations in head pose and lightning. Videos in this database are annotated with valence and arousal values for each frame. A total of 254 videos of this database are selected for training and the remaining 46 videos were used for evaluating the participants in the challenge. Since the evaluation set is not publicly available, we selected 26 sequences of the training set as our validation set (in a subject-independent manner). Figure~\ref{fig:aff-in-wildExamples} provides a few examples from Affect-in-Wild database.

\subsection{Evaluation Metrics for Dimensional Model}\label{sec:evalmetrics}

In order to evaluate the methods in the dimensional model of affect, we calculate and report Root Mean Square Error (RMSE), Correlation Coefficient (CC), Concordance Correlation Coefficient (CCC), and sign agreement (SAGR) metrics for the methods. In the following, we briefly explain the definitions of these metrics.

\vspace{10pt}
\noindent\textbf{Root Mean Square Error (RMSE)} is the most common evaluation metric in a continuous domain which is
defined as:
\begin{equation}
\text{RMSE} = \sqrt{\frac{1}{n} \sum_{i=1}^{n}\left(\hat{\theta}_i-\theta_i\right)^2}
\end{equation}
where $\hat{\theta}_i$ and $\theta_i$ are the prediction and the ground-truth of $i^{\text{th}}$ sample, and $n$ is the number of samples. RMSE-based evaluation metrics can heavily weigh the outliers~\cite{bermejo2001oriented}, and they do not consider covariance of the data. 

\vspace{10pt}
\noindent\textbf{Pearson's Correlation Coefficient (CC)} overcomes RMSE's reliance on outliers~\cite{nicolaou2011continuous} and it is defined as: 
\begin{equation}
\text{CC} = \frac{\text{COV}\{\hat{\theta}, \theta\}}{\sigma_{\hat{\theta}}\sigma_{\theta}} = 
\frac{E \left[(\hat{\theta}-\mu_{\hat{\theta}})(\theta-\mu_{\theta})\right]}{\sigma_{\hat{\theta}}\sigma_{\theta}}\label{eq:CC}
\end{equation}
where COV is covariance function.

\vspace{10pt}
\noindent\textbf{Concordance Correlation Coefficient (CCC)} is another metric~\cite{valstar2016avec} and combines CC with the square difference between the means of two compared time series:

\begin{equation}
\rho_c = \frac{2\rho \sigma_{\hat{\theta}} \sigma_{\theta}}{\sigma_{\hat{\theta}}^2 + \sigma_{\theta}^2 + (\mu_{\hat{\theta}} - \mu_\theta)^2}\label{eq:CCC}
\end{equation}
where $\rho$ is the Pearson correlation coefficient (CC) between two time-series (\eg, prediction and ground-truth), $\sigma_{\hat{\theta}}^2$ and $\sigma_{\theta}^2$ are the variance of each time series, $\sigma_{\hat{\theta}}$ and $\sigma_{\theta}$ are the standard deviation of each, and $\mu_{\hat{\theta}}$ and $\mu_{\theta}$ are the mean value of each. Unlike CC, the predictions that are well correlated with the ground-truth but shifted in values are penalized in proportion to the deviation in the CCC.

\begin{table*}[htbp]
\centering
\caption{Number of parameters and FLOPs for different depths of BReG-NeXt}
\label{tab:differentDepthsParameters}
\begin{tabular}{lc|c|c|c|c|c|c|c}
	\cline{2-9}
	& 26-layer         & 32-layer         & 38-layer         & 44-layer         & 50-layer         & 56-layer         & 62-layer         & 68-layer         \\ \hline\hline
	\multicolumn{1}{l|}{\begin{tabular}[c]{@{}l@{}}number of\\ parameters\end{tabular}} & 1.5M             & 1.9M             & 2.3M             & 2.6M             & 3.1M             & 3.4M             & 3.8M             & 4.2M             \\ \hline
	\multicolumn{1}{l|}{FLOPs}                                                          & $0.76\times10^7$ & $0.95\times10^7$ & $1.15\times10^7$ & $1.34\times10^7$ & $1.53\times10^7$ & $1.73\times10^7$ & $1.92\times10^7$ & $2.12\times10^7$ \\ \hline
\end{tabular}
\end{table*}
\begin{table}[tbp]
	\centering
	\caption{Number of annotated images for each expression on the studied databases }
	\label{tab:NumImages}
	\begin{tabular}{l|c|c}
		\hline
		\multicolumn{1}{c|}{\textbf{Expression}} & \textbf{AffectNet}& \textbf{FER2013} \\ \hline \hline
		Neutral                                   & 80,276     & 6,198     \\ \hline
		Happy                                     & 146,198    &  8,989    \\ \hline
		Sad                                       & 29,487    &  6,077     \\ \hline
		Surprise                                  & 16,288    &    4,002   \\ \hline
		Fear                                      & 8,191     &  5,121     \\ \hline
		Disgust                                   & 5,264     &   547    \\ \hline
		Anger                                     & 28,130    &   4,953    \\ \hline
		Contempt                                  & 5,135     &     -  \\ \hline
	\end{tabular}
\end{table}

\vspace{10pt}
\noindent\textbf{SAGR:} The value of valence and arousal fall within the interval of [-1,+1] and correctly predicting their signs are essential in many emotion-prediction applications. Therefore, we use Sign AGReement (SAGR) metric as proposed in~\cite{nicolaou2011continuous} to evaluate the performance of a valence and arousal prediction system with respect to the sign agreement. SAGR is defined as:
\begin{equation}
\text{SAGR} = \frac{1}{n}\sum_{i=1}^{n}\delta{(\text{sign}(\hat{\theta}_i),\text{sign}(\theta_i))}
\end{equation}
where $\delta$ is the Kronecker delta function, defined as:
\begin{equation}
\delta{(a,b)} = 
\begin{cases}
1,&				 a = b\\
0,&              a \neq b
\end{cases}
\end{equation}

\subsection{Results}

\begin{figure}[t]
	\centering
	\subfigure[Categorical]{\label{fig:depthsErrorcat}\includegraphics[width=.45\linewidth]{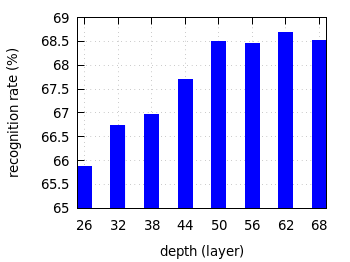}}
	\qquad
	\subfigure[Dimensional]{\label{fig:depthsErrordim}\includegraphics[width=.45\linewidth]{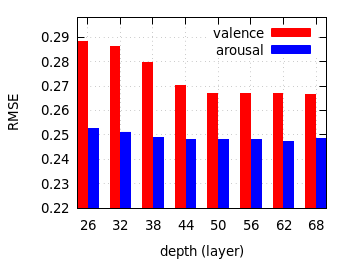}}
	\caption{Result of experimenting different depths for BReG-NeXt on categorical and dimensional models of AffectNet database}\label{fig:depthsError}
\end{figure}

As mentioned before, we have selected BReG-NeXt-50 as our final architecture. Similar to ResNet, by increasing the number of layers, the recognition rate increases in BReG-NeXt. However, considering the trade-off between FLOPs and recognition rate, after a certain point, recognition rate plateaus and it is not efficient to increase the depth of the network anymore. Table~\ref{tab:differentDepthsParameters} shows the number of training parameters as well as the number of FLOPs for different depths of BReG-NeXt and Figure~\ref{fig:depthsError} shows recognition rate and RMSE of AffectNet database on the validation set for the categorical and dimensional models of affect, respectively. It can be seen that the number of training parameters and FLOPs linearly increase with adding more depth to the network while the recognition rate's improvement is negligible after BReG-NeXt-50. For instance, the highest recognition rate in the categorical model is achieved by BReG-NeXt-62 (68.69\%) which shows only 0.19\% improvement over BReG-NeXt-50 (68.50\%). This is while the number of training parameters and FLOPs for BReG-NeXt-62 are more than those of BReG-NeXt-50 by 22\% and 25\%, respectively (Figure~\ref{fig:depthsErrorcat}). Similar behavior in dimensional model happens where by increasing the depth of the network, RMSE of the validation set plateaus (Figure~\ref{fig:depthsErrordim}). This phenomenon is more visible for the valence predictions than it is for arousal's but it occurs for both dimensions after a certain point. We need to mention that for each depth increment we add one  unit to the residual unit ($F_1$ in Figure~\ref{fig:overview}). Therefore, three units are added in each increment and since there are two convolution layers in each residual unit, thus the number of layers added in each increment is six. Based on this argument we decided to choose BReG-NeXt-50 as our final proposed networks for our method.

\begin{table*}[th]
	\centering
	\caption{Recognition rates (\%) in categorical model of affect}
	\label{tab:categoricalresults}
	\begin{tabular}{lc|c|c|c|c|c|c}
		\cline{2-8}
		& \multicolumn{1}{l|}{\textbf{ResNet-32}} & \textbf{ResNet-50} & \textbf{BReG-Net-32} & \textbf{BReG-Net-50} & \textbf{BReG-NeXt-32} & \textbf{BReG-NeXt-50} & \textbf{\begin{tabular}[c]{@{}c@{}}state-of-the-art\\ methods\end{tabular}} \\ \hline\hline
		\multicolumn{1}{l|}{\textbf{AffectNet}} & 59.45   &  63.33      &  65.66    & 66.96 & 66.74    & \textbf{68.50}        &  \begin{tabular}[c]{@{}c@{}}58.0~\cite{mollahosseini2017affectnet}, 57.31~\cite{zeng2018facial} \\  48~\cite{wang2018two}, 62.11~\cite{8643924}\\ 58~\cite{hewitt2018cnn}, 60~\cite{kollias2018generating} \\ 61.5~\cite{chen2019facial} \end{tabular}                                                                          \\ \hline
		\multicolumn{1}{l|}{\textbf{FER2013}}   & 65.81 & 67.15       &  67.86    & 69.21 & 69.11    & \textbf{71.53}        &  \begin{tabular}[c]{@{}c@{}}69.3~\cite{tang2013deep},  66.4~\cite{c4} \\ 71.2~\cite{vielzeuf2017temporal} \end{tabular}                                                                          \\ \hline
	\end{tabular}
\end{table*}

\begin{table*}[htbp]
\centering
\caption{Evaluation metrics on BReG-NeXt-50 for categorical model of affect}
\label{tab:prec-rec-f1}
\begin{tabular}{llccccccccc}
\cline{3-11}
                                                &                                & \multicolumn{1}{l|}{\textbf{neutral}} & \multicolumn{1}{l|}{\textbf{happy}} & \multicolumn{1}{l|}{\textbf{sad}} & \multicolumn{1}{l|}{\textbf{surprise}} & \multicolumn{1}{l|}{\textbf{fear}} & \multicolumn{1}{l|}{\textbf{disgust}} & \multicolumn{1}{l|}{\textbf{angry}} & \multicolumn{1}{l||}{\textbf{contempt}} & \textbf{average} \\ \hline\hline
\multicolumn{1}{l|}{\multirow{3}{*}{\textbf{AffectNet}}} & \multicolumn{1}{l|}{precision} & \multicolumn{1}{c|}{0.72}        & \multicolumn{1}{c|}{0.78}        & \multicolumn{1}{c|}{0.62}        & \multicolumn{1}{c|}{0.66}        & \multicolumn{1}{c|}{0.67}        & \multicolumn{1}{c|}{0.62}        & \multicolumn{1}{c|}{0.63}        & \multicolumn{1}{c||}{0.67}        & 0.69             \\
\multicolumn{1}{l|}{}                           & \multicolumn{1}{l|}{recall}    & \multicolumn{1}{c|}{0.53}        & \multicolumn{1}{c|}{0.89}        & \multicolumn{1}{c|}{0.66}        & \multicolumn{1}{c|}{0.74}        & \multicolumn{1}{c|}{0.63}        & \multicolumn{1}{c|}{0.77}        & \multicolumn{1}{c|}{0.61}        & \multicolumn{1}{c||}{0.58}        & 0.69             \\
\multicolumn{1}{l|}{}                           & \multicolumn{1}{l|}{F1-score}  & \multicolumn{1}{c|}{0.61}        & \multicolumn{1}{c|}{0.83}        & \multicolumn{1}{c|}{0.64}        & \multicolumn{1}{c|}{0.70}        & \multicolumn{1}{c|}{0.65}        & \multicolumn{1}{c|}{0.69}        & \multicolumn{1}{c|}{0.62}        & \multicolumn{1}{c||}{0.62}        & 0.68             \\ \hline\hline
\multicolumn{1}{l|}{\multirow{3}{*}{\textbf{FER2013}}}   & \multicolumn{1}{l|}{precision} & \multicolumn{1}{c|}{0.69}        & \multicolumn{1}{c|}{0.88}        & \multicolumn{1}{c|}{0.59}        & \multicolumn{1}{c|}{0.78}        & \multicolumn{1}{c|}{0.52}        & \multicolumn{1}{c|}{0.62}        & \multicolumn{1}{c|}{0.60}        & \multicolumn{1}{c||}{-}           & 0.71             \\
\multicolumn{1}{l|}{}                           & \multicolumn{1}{l|}{recall}    & \multicolumn{1}{c|}{0.69}        & \multicolumn{1}{c|}{0.90}        & \multicolumn{1}{c|}{0.62}        & \multicolumn{1}{c|}{0.80}        & \multicolumn{1}{c|}{0.46}        & \multicolumn{1}{c|}{0.12}        & \multicolumn{1}{c|}{0.65}        & \multicolumn{1}{c||}{-}           & 0.72             \\
\multicolumn{1}{l|}{}                           & \multicolumn{1}{l|}{F1-score}  & \multicolumn{1}{c|}{0.68}        & \multicolumn{1}{c|}{0.89}        & \multicolumn{1}{c|}{0.60}        & \multicolumn{1}{c|}{0.79}        & \multicolumn{1}{c|}{0.49}        & \multicolumn{1}{c|}{0.21}        & \multicolumn{1}{c|}{0.62}        & \multicolumn{1}{c||}{-}           & 0.71             \\ \hline
                                                                                                                                                                                                   
\end{tabular}
\end{table*}

\vspace{10pt}
\noindent\textbf{Categorical Model:} Facial expression databases are usually highly skewed. This form of imbalanced data is referred to as ``intrinsic variation", \ie, it is a direct result of the nature of expressions in the real world. Therefore, this phenomena occurs in both categorical and dimensional models of affect. For example, Caridakis \textit{et al.}~\cite{caridakis2008user} reported that a bias toward the first quadrant of valence/arousal circumplex (positive arousal, positive valence) exists in the SAL database. Table~\ref{tab:NumImages} shows an imbalanced number of images for each emotion category in AffectNet and FER2013 databases used in this work. We face two problems while working with imbalanced data. First, training with an imbalanced distribution often causes the learning algorithms to perform poorly on the less-represented classes~\cite{he2009learning}. Second, the imbalance in the test/validation data distribution can affect the performance metrics of the methods significantly.

Jeni \etal~\cite{jeni2013facing} showed that with exception of the area under the ROC curve (AUC), all other studied evaluation metrics, \ie, Accuracy, F1-score, Cohen's kappa~\cite{cohen1960}, Krippendorf's Alpha~\cite{krippendorff1970estimating}, and area under Precision-Recall curve (AUC-PR) are affected by skewed distributions dramatically. While AUC is unaffected by skew, precision-recall curves suggested that AUC may mask poor performance. There have been some attempts to overcome this problem. In~\cite{mollahosseini2017affectnet,hasani2019bounded} weighted loss functions are used in which the loss function heavily penalizes the networks for misclassifying examples from
under-represented classes while penalizing networks less for misclassifying examples from well-represented classes.

Recently, \textit{focal loss}~\cite{lin2017focal} has drawn attention for imbalanced data training. Focal loss is the reshaping of cross entropy loss such that it down-weights the loss assigned to well-classified examples. Focal loss focuses on training on a sparse set of hard examples and prevents the vast number of easy negatives from overwhelming the network during training. Formally, for binary classification, cross entropy loss is defined as $\text{CE}(p_t) = -\log(p_t)$ where $p_t$ is defined as:
\begin{equation}
p_t = 
\begin{cases}
p&				 \text{if}~y = 1\\
1-p&              \text{otherwise}
\end{cases}
\end{equation}
in which $y \in \{\pm 1\}$ and $p \in [0,1]$ is the model's estimated probability for the class $y = 1$. In focal loss, modulating factor $(1 - p_t)^\gamma$, and balancing factor $\alpha_t$ is multiplied to the cross entropy loss as follows:
\begin{equation}
\text{FL}(p_t) = - \alpha_t(1-p_t)^\gamma\log(p_t)
\end{equation}
where $\gamma \geq  0$ is called \textit{focusing parameter}. In our experiments on the categorical model of affect, we use focal loss (with $\alpha_t = 0.25$ and $\gamma = 2$) as our loss function for the optimizer. 

Table~\ref{tab:categoricalresults} shows the result of our experiments in the categorical model of affect on AffectNet and FER2013 databases. All of the reported numbers are the results of our experiments on the validation set of these databases. As can be seen, our proposed modification of the ResNet module achieves better recognition rates compared to their counterpart on original ResNet. Our method also outperforms the existing methods on both AffectNet and FER2013 databases. For state-of-the-art methods mentioned in Table~\ref{tab:categoricalresults}, Mollahosseini \emph{et al.}~\cite{mollahosseini2017affectnet} uses AlexNet, Wiles \emph{et al.}~\cite{wiles2018self} achieved 74.4 for AUC, and~\cite{c4} uses an Inception-based method to classify the expressions,~\cite{tang2013deep} trained deep learning methods combined with SVMs, in~\cite{wang2018two} a two-level attention
with two-stage multi-task learning framework is proposed for facial emotion
estimation, and in~\cite{vielzeuf2017temporal} a multi-modal approach for video emotion classification is used by combining VGG and C3D as image descriptors. It is worth to mention that our proposed method is considerably shallower than many of the methods proposed in the field.

\begin{figure}[tbp]
	\centering
	\subfigure[AffectNet]{\label{fig:AffectNetCM}\includegraphics[width=.43\linewidth]{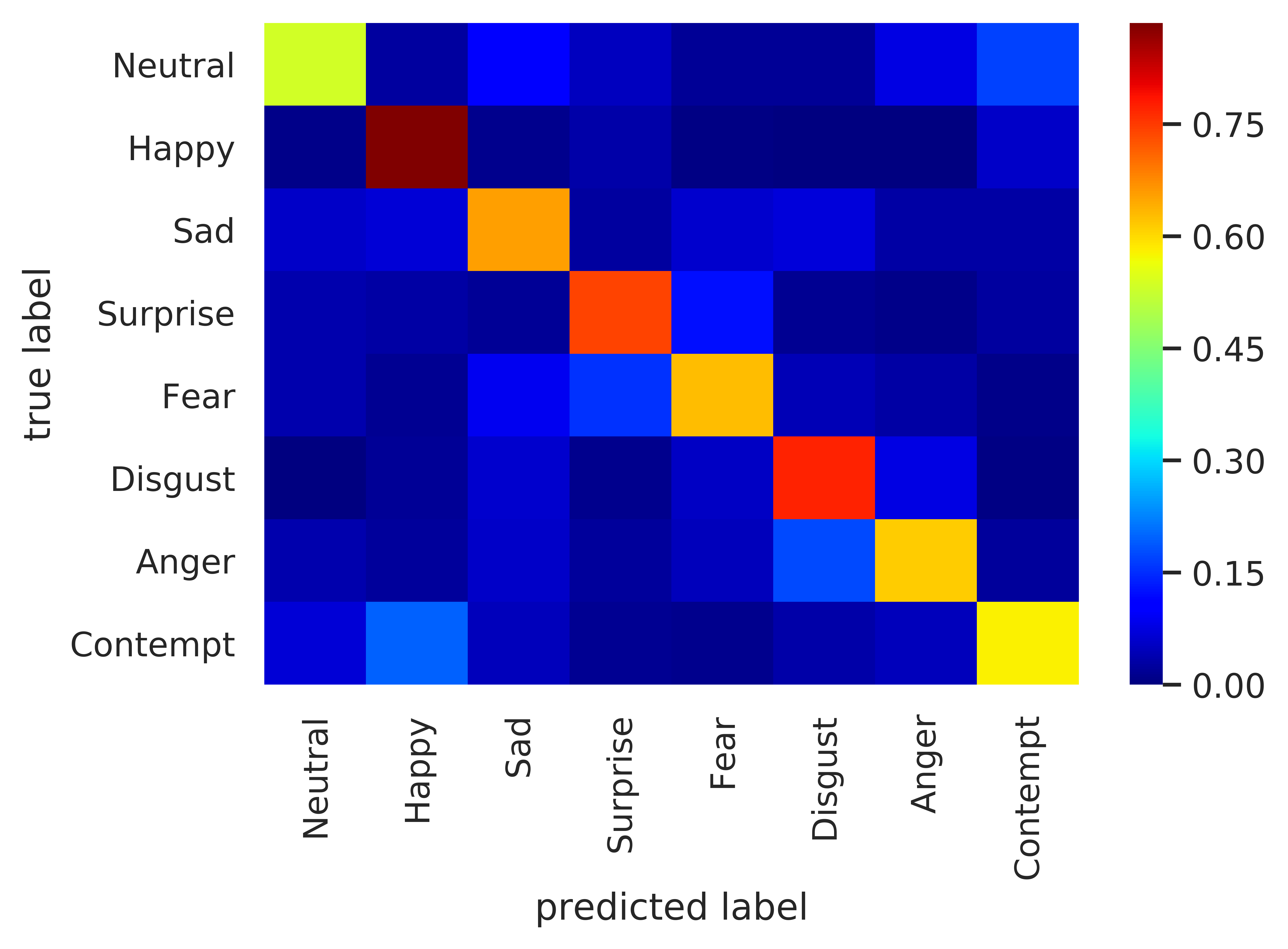}}
	\qquad
	\subfigure[FER2013]{\label{fig:FER2013CM}\includegraphics[width=.43\linewidth]{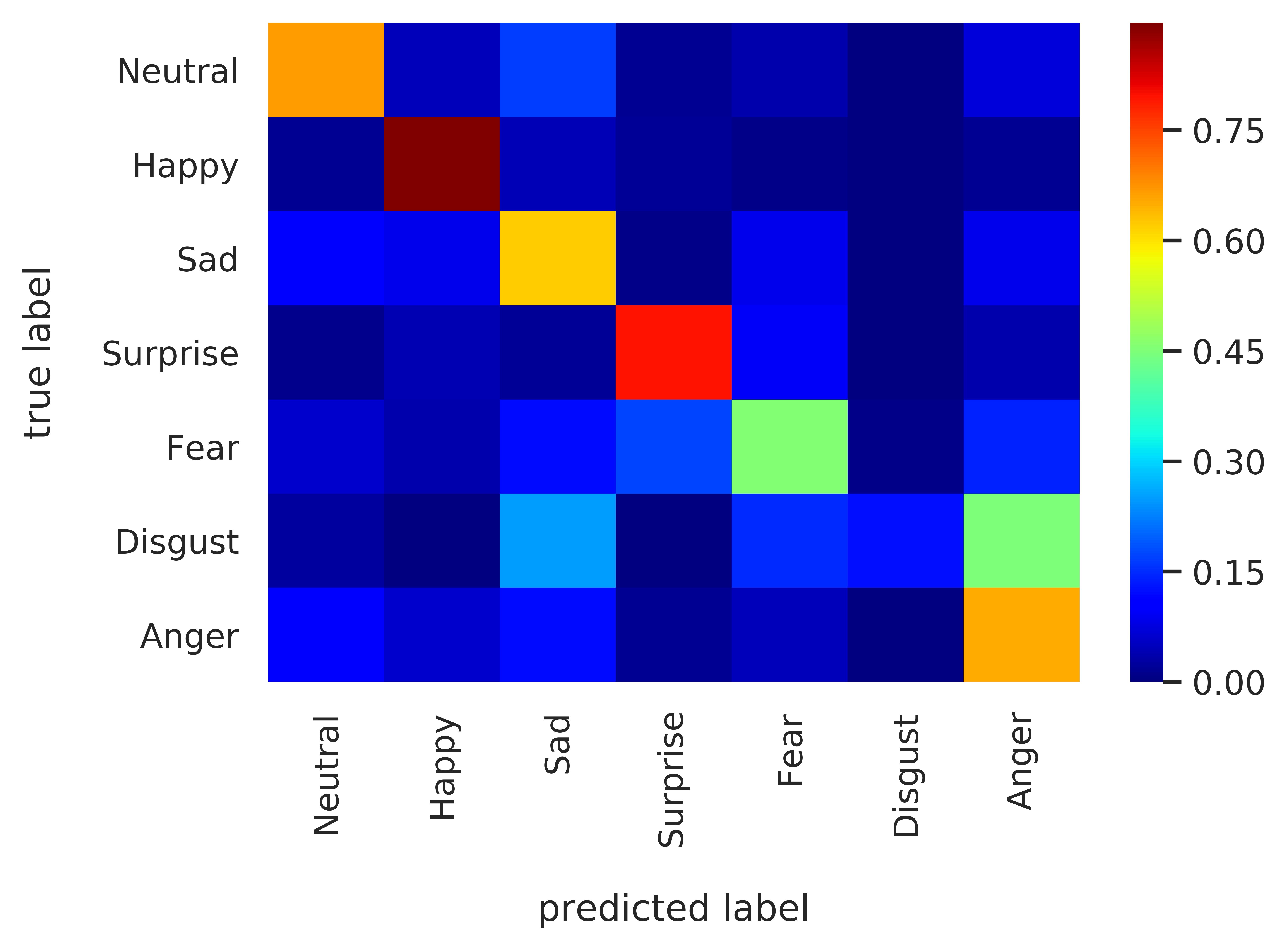}}
	\caption{Confusion matrix of BReG-NeXt-50 on AffectNet (a) and FER2013 (b) on categorical model of affect}\label{fig:confusionMatrix}
\end{figure}

Table~\ref{tab:prec-rec-f1} provides additional evaluation metrics on BReG-NeXt-50 for the categorical model of affect. It can be seen that in both databases ``Happy" is recognized more accurately compared to other emotions. This is because ``Happy" has a considerable number of samples in both training sets and also it is considerably distinguishable from other emotions in its shape nature. On AffectNet, the rest of the emotions have close recognition rate (in terms of F1-score) to the average which shows that BReG-NeXt-50 is not excessively biased towards any emotion. On FER2013, however, ``disgust" has a lower recognition rate comparing to the other emotions. This is due to the fact that this category is barely represented in the dataset (only 1.5\% of the entire data based on Table~\ref{tab:NumImages}) to the extent that  focal loss is not able to assign enough priority for the samples of this category.

\begin{figure*}[th]
	\centering
	\subfigure[AffectNet]{\label{fig:AffectNetResutls}\includegraphics[width=.75\linewidth]{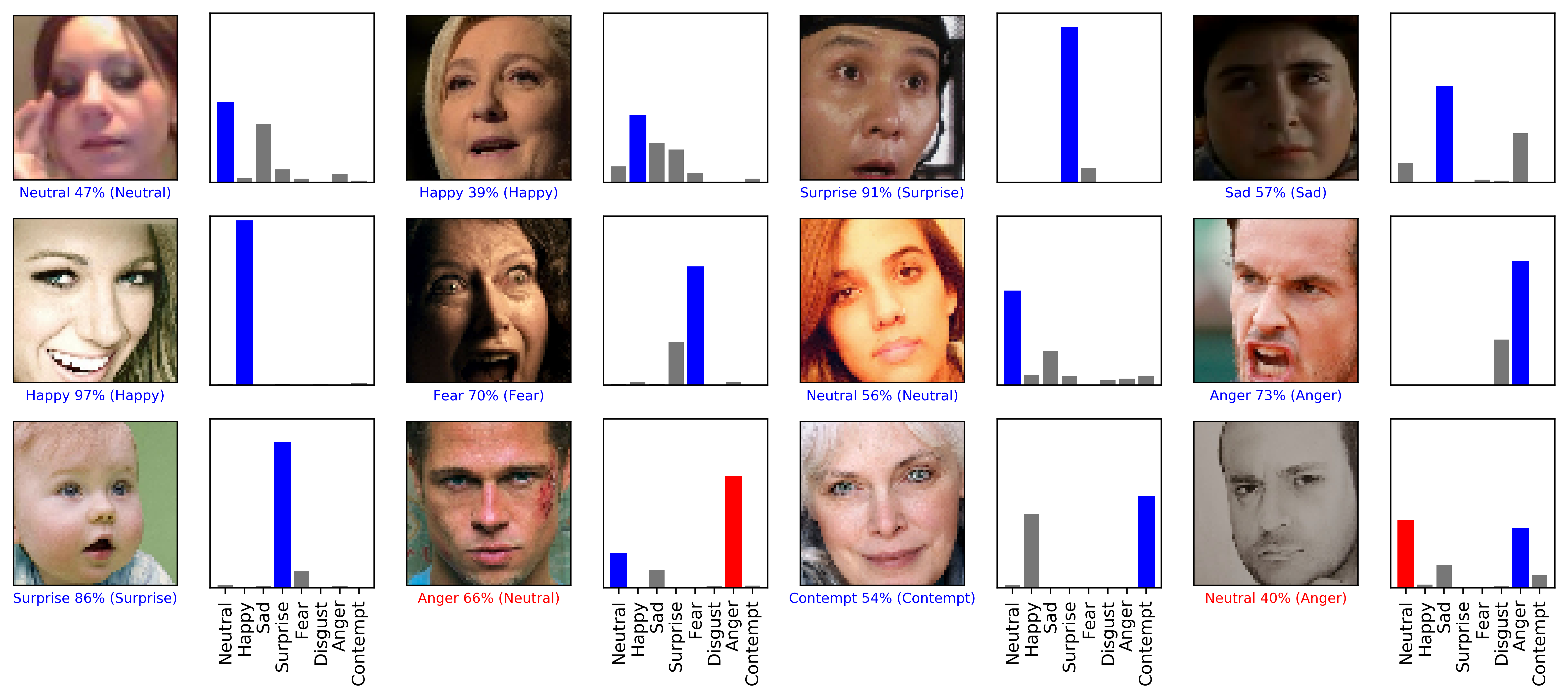}} \\
	\subfigure[FER2013]{\label{fig:FER2013Results}\includegraphics[width=.75\linewidth]{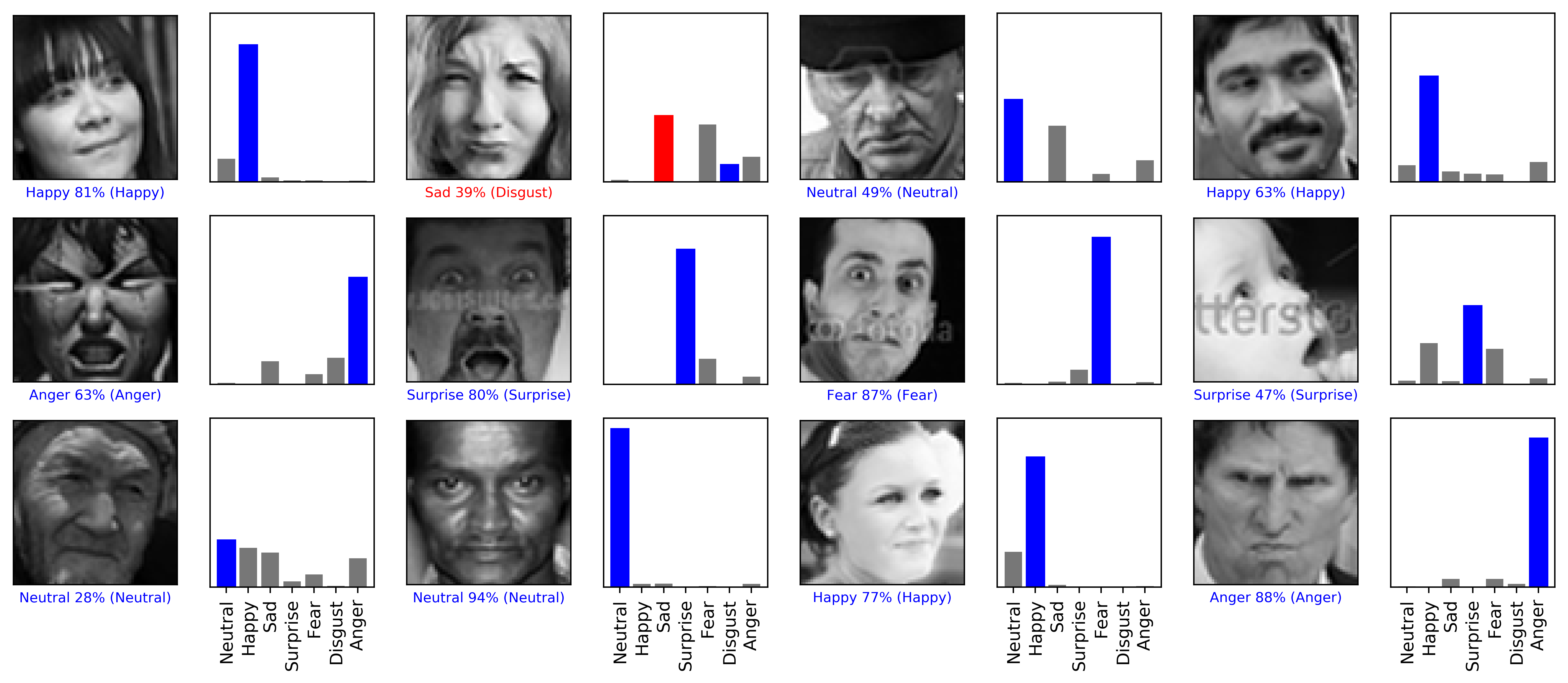}}
	\caption{Example predictions of BReG-NeXt-50 on AffectNet (a) and FER2013 (b). The text below images indicates predicted label, confidence, and true label of the image, receptively. Blue indicates correct classification while red shows misclassification.}\label{fig:predictions}

\end{figure*}

\begin{figure*}[th]
	\centering
		\subfigure[Input image]{\includegraphics[width=.13\linewidth]{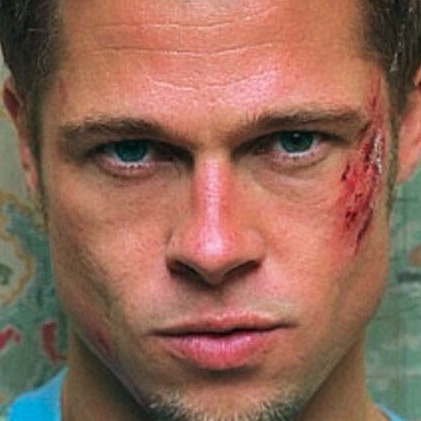}}
		\qquad
		\subfigure[BReG-NeXt@15]{\includegraphics[width=.135\linewidth]{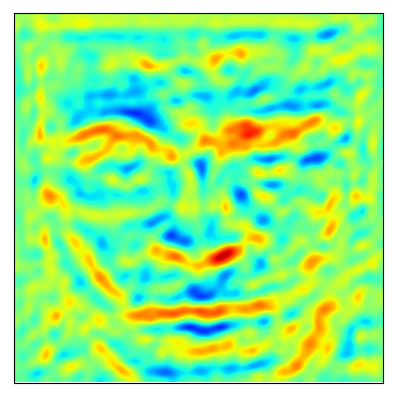}}
		\qquad
		\subfigure[ResNet@15]{\includegraphics[width=.135\linewidth]{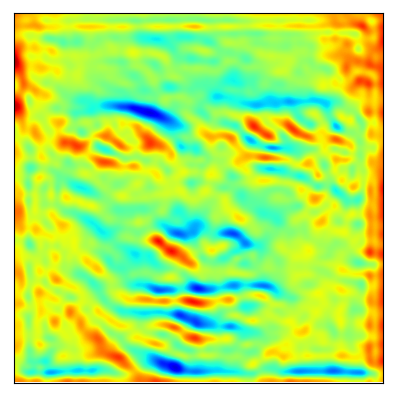}}
		\qquad
		\subfigure[BReG-NeXt@17]{\includegraphics[width=.135\linewidth]{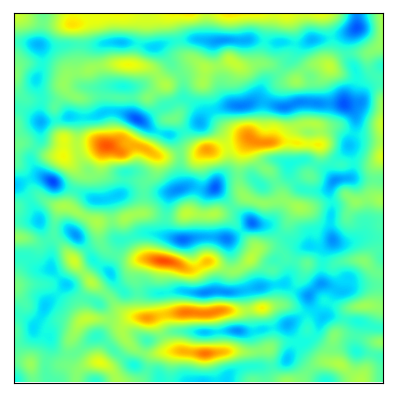}}
		\qquad
		\subfigure[ResNet@17]{\includegraphics[width=.135\linewidth]{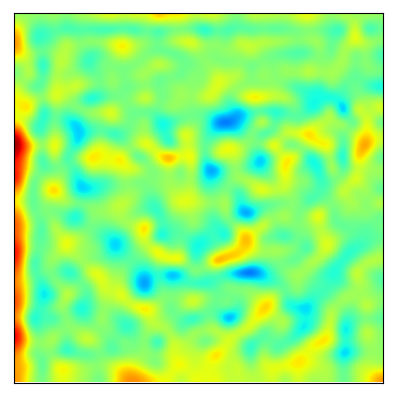}}\\
		\qquad
		\subfigure[BReG-NeXt@33]{\includegraphics[width=.135\linewidth]{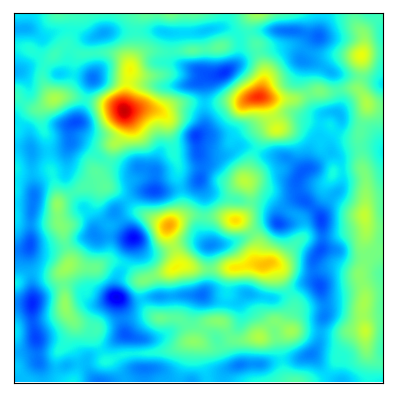}}
		\qquad
		\subfigure[ResNet@33]{\includegraphics[width=.135\linewidth]{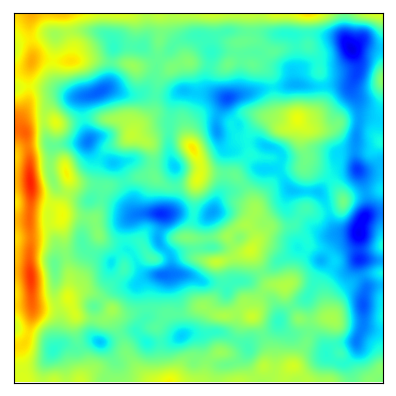}}
		\qquad
		\subfigure[BReG-NeXt@50]{\includegraphics[width=.135\linewidth]{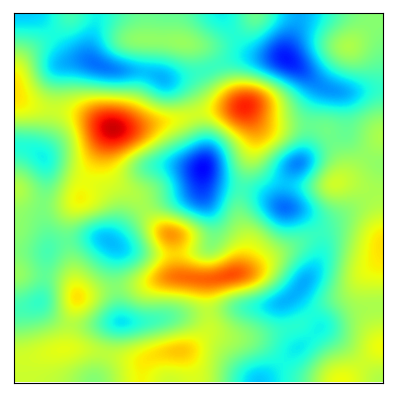}}
		\qquad
		\subfigure[ResNet@50]{\includegraphics[width=.135\linewidth]{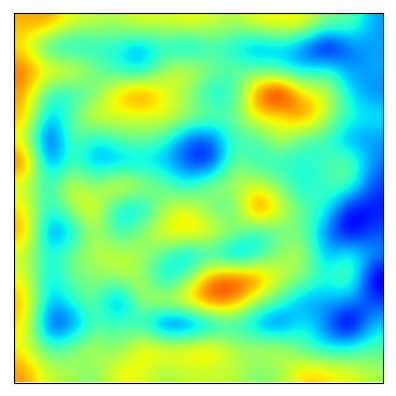}}
		\caption{An example input image (a) and its corresponding feature map at different depths of BReG-NeXt-50 and ResNet-50. Figures (b), (d), (f), and (h) show examples of resulting feature map at the depths 15, 17, 33, and 50 in BReG-NeXt-50, respectively. Figures (c), (e), (g), and (i) show examples of resulting feature map at the depths 15, 17, 33, and 50 in ResNet-50, respectively.}\label{fig:featuremaps}
\end{figure*}

Figure~\ref{fig:confusionMatrix} shows the confusion matrix of BReG-NeXt-50 on the categorical model. On AffectNet, the most confusion occurs between ``Happy" and ``Contempt" which is not unexpected as these two categories are very similar to the extent that distinguishing between the two is difficult even for humans. On FER2013, as mentioned earlier, the low number of samples for ``Disgust" has caused the main confusion for the network. Other categories, however, have been distinguished well considering the fact that the database is very challenging.

\begin{table*}[htbp]
	\centering
	\caption{RMSE values of our experiments on dimensional model of affect}
	\label{tab:dimensionalresults}
	\resizebox{\textwidth}{!}{
	\begin{tabular}{lcc|c|c|c|c|c|c}
		\cline{3-9}
		& \multicolumn{1}{l}{}                & \textbf{ResNet-32} & \textbf{ResNet-50} & \textbf{BReG-Net-32}& \textbf{BReG-Net-50} & \textbf{BReG-NeXt-32} & \textbf{BReG-NeXt-50} & \textbf{\begin{tabular}[c]{@{}c@{}}state-of-the-art\\ methods\end{tabular}} \\ \hline\hline
		\multicolumn{1}{l|}{\multirow{3}{*}{\textbf{AffectNet}}}      & \multicolumn{1}{c|}{valence}        & 0.2888            & 0.2811             & 0.2676 & \textbf{0.2555} & 0.2863                & 0.2668                &\begin{tabular}[c]{@{}c@{}} 0.37~\cite{mollahosseini2017affectnet}, 0.4406~\cite{JANG201917} \\ 0.444~\cite{langholz2019oculum}, 0.353~\cite{wang2018two}\\     \end{tabular}                                                 \\ \cline{2-9}
		\multicolumn{1}{l|}{}                                         & \multicolumn{1}{c|}{arousal}        & 0.3376             & 0.3221              & 0.2970 & 0.2852 & 0.2492                & \textbf{0.2482}       &\begin{tabular}[c]{@{}c@{}} 0.41~\cite{mollahosseini2017affectnet},  0.3937~\cite{JANG201917}\\ 0.389~\cite{langholz2019oculum}, 0.364~\cite{wang2018two}\\     \end{tabular}                                                            \\ \cline{2-9} 
		\multicolumn{1}{l|}{}                                         & \multicolumn{1}{c|}{\textbf{total}} & 0.3142             & 0.3023            & 0.2826 & 0.2708 & 0.2684                & \textbf{0.2577}       & \begin{tabular}[c]{@{}c@{}} 0.3905~\cite{mollahosseini2017affectnet}, 0.359~\cite{wang2018two}\\     \end{tabular}                                                         \\ \hline\hline
		\multicolumn{1}{l|}{\multirow{3}{*}{\textbf{Affect-in-Wild}}} & \multicolumn{1}{c|}{valence}        & 0.3023             & 0.2768              & 0.2855 & 0.2680 & 0.2873             & \textbf{0.2644}           & 0.27~\cite{hasani2017facial_dimensional}                                                        \\\cline{2-9} 
		\multicolumn{1}{l|}{}                                         & \multicolumn{1}{c|}{arousal}        & 0.3450             & 0.3448             & 0.3351 & 0.3180 & 0.3119             & \textbf{0.3102}       & 0.36~\cite{hasani2017facial_dimensional}                                                                \\ \cline{2-9} 
		\multicolumn{1}{l|}{}                                         & \multicolumn{1}{c|}{\textbf{total}} & 0.3244             & 0.3127              & 0.3113 & 0.2941 & 0.2950                & \textbf{0.2882}       & 0.3182~\cite{hasani2017facial_dimensional}                                                               \\ \hline	
	\end{tabular}
}
\end{table*}

Figure~\ref{fig:predictions} depicts a few examples of predictions made by BReG-NeXt-50 with their corresponding confidence score. It can be seen that our method performs well in predicting most of the instances and for misclassified examples, networks predictions are so a certain degree relevant to the input pictures. Also, our method performs well specifically on the difficult categories of neutral and contempt; it is able to recognize the subtle facial properties for these challenging categories.

Figure~\ref{fig:featuremaps} shows an example input image and its corresponding feature map at different depths of BReG-NeXt-50 and ResNet-50. It can be seen that at each layer, BReG-NeXt-50 performs considerably better in distinguishing important components of the face such as eyes and mouth which results in better recognition at the end.

\begin{table*}[htbp]
	\centering
	\caption{Evaluation metrics on BReG-NeXt-50 for dimensional model of affect}
	\label{tab:cc-ccc-sagr}
\begin{tabular}{l|cc|cc|cc|}
\cline{2-7}
                                              & \multicolumn{2}{c|}{\textbf{CC}} & \multicolumn{2}{c|}{\textbf{CCC}} & \multicolumn{2}{c|}{\textbf{SAGR}} \\ \cline{2-7} 
                                              & valence         & arousal        & valence         & arousal         & valence          & arousal         \\ \hline
\multicolumn{1}{|l|}{\textbf{AffectNet}}      & 0.78             &  0.86         &  0.74          & 0.85             & 0.77             &  0.82           \\ \hline
\multicolumn{1}{|l|}{\textbf{Affect-in-Wild}} & 0.42            & 0.40           & 0.37            & 0.31            & 0.77              &  0.75          \\ \hline
\end{tabular}
\end{table*}

\begin{figure}[th]
	\centering
	\subfigure[ AffectNet valence error histogram]{\label{fig:AffectNetErrorHistVal}\includegraphics[width=.45\linewidth]{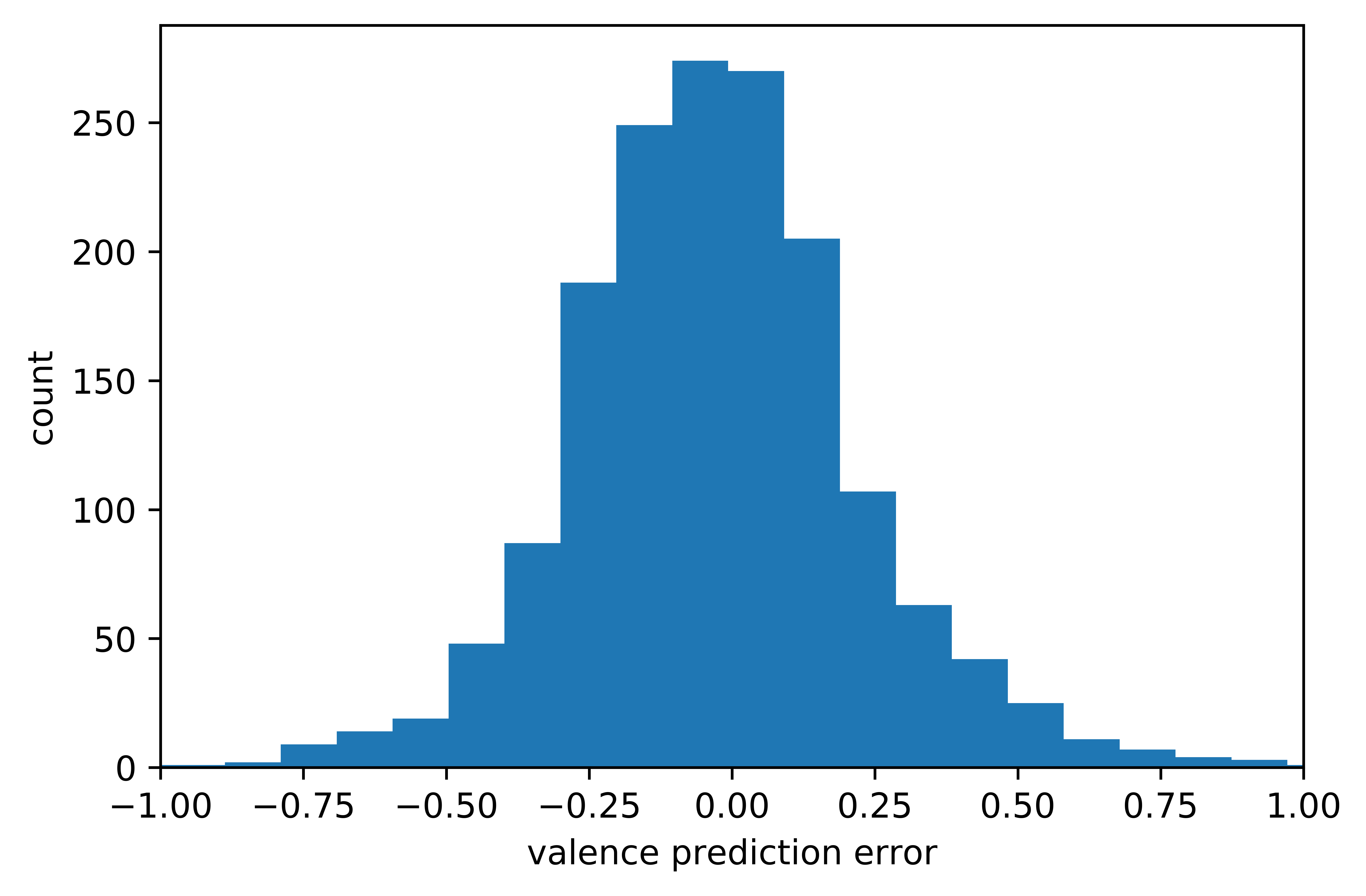}}
	\qquad
	\subfigure [AffectNet arousal error histogram]{\label{fig:AffectNetErrorHistArousal}\includegraphics[width=.45\linewidth]{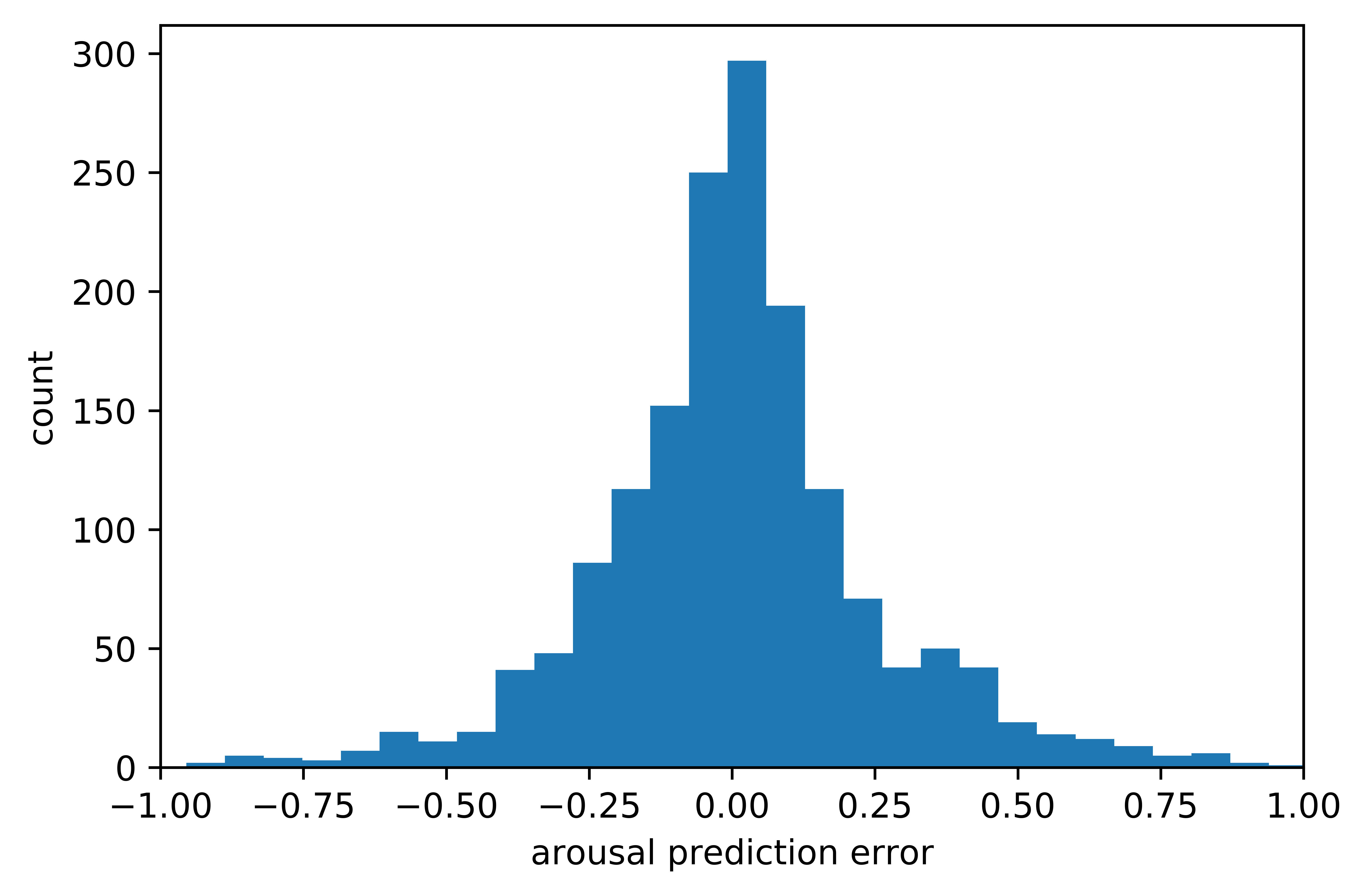}}\\
	\subfigure[Affect-in-Wild valence error histogram]{\label{fig:AffErrorHistVal}\includegraphics[width=.45\linewidth]{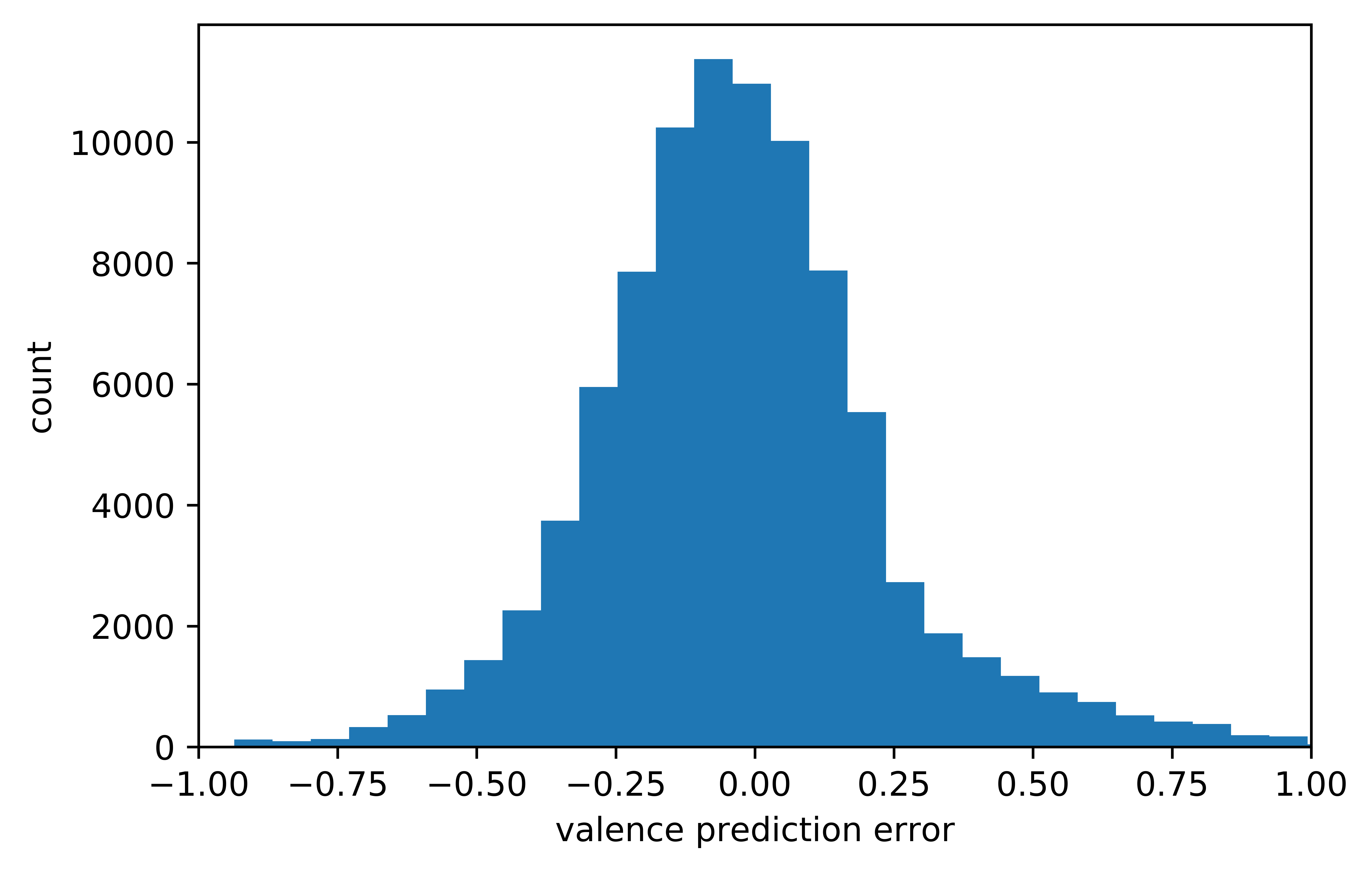}}
	\qquad
	\subfigure[Affect-in-Wild arousal histogram]{\label{fig:AffErrorHistArousal}\includegraphics[width=.45\linewidth]{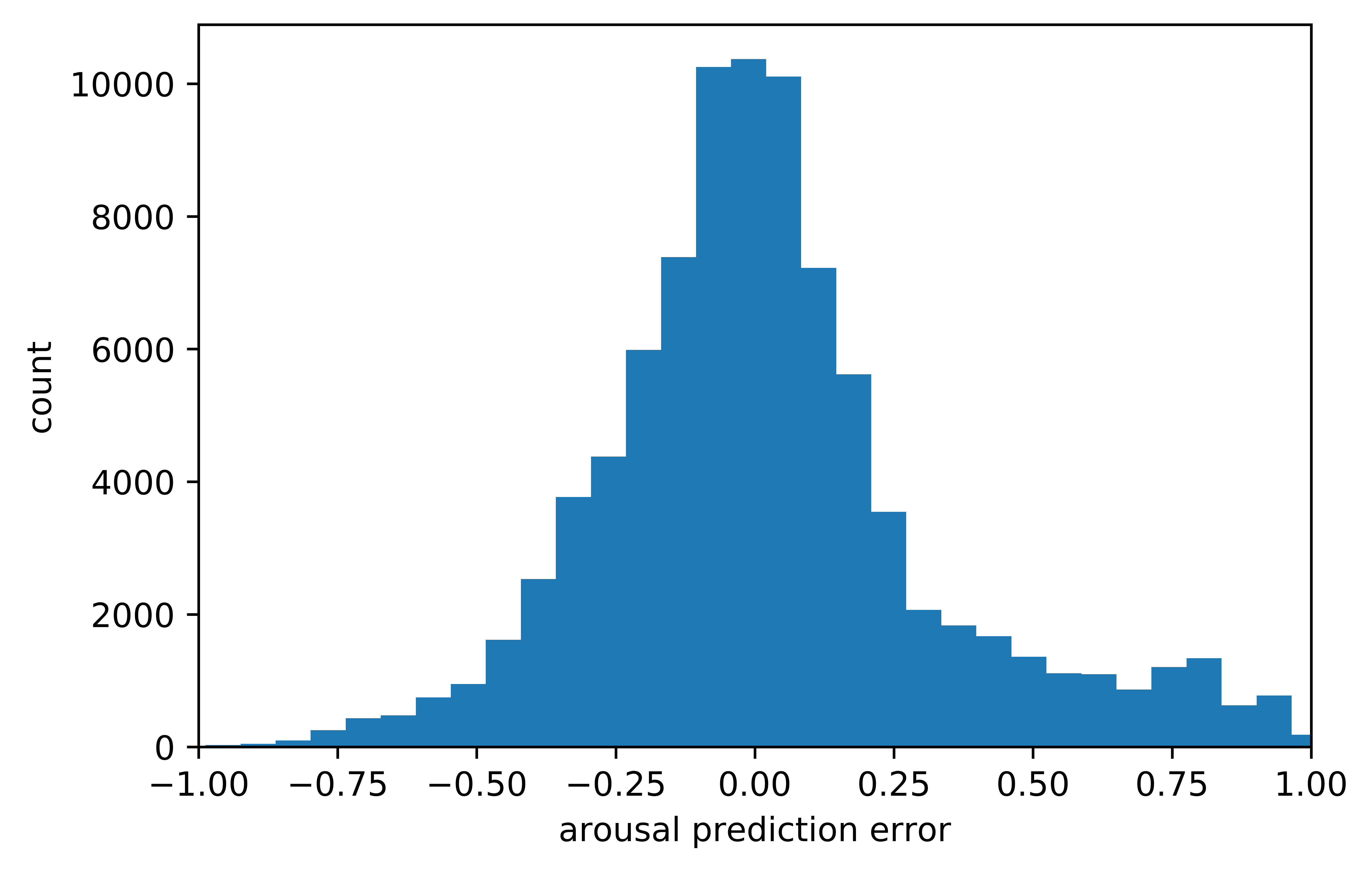}}
	\caption{Error histogram of BReG-NeXt-50 on AffectNet (a and b), and Affect-in-Wild (c and d) databases on dimensional model of affect }\label{fig:errorHist}
\end{figure}

\vspace{10pt}
\noindent\textbf{Dimensional Model:} Table~\ref{tab:dimensionalresults} shows the results of our experiments in the dimensional model of affect on the validation set of the AffectNet and Affect-in-Wild databases. Same as the categorical model of affect, our method achieves better results (lower loss and RMSE) in total (both valence and arousal considered together) compared to their corresponding network on ResNet and BReG-Net. For state-of-the-art methods mentioned in Table~\ref{tab:dimensionalresults}, Mollahosseini~\etal~\cite{mollahosseini2017affectnet} use AlexNet, and~\cite{hasani2017facial_dimensional} uses an Inception-ResNet-based method to classify the expressions. In~\cite{wang2018two} a two-level attention
with two-stage multi-task learning framework is proposed for facial emotion
estimation on static images using Bi-directional Recurrent Neural Networks (Bi-RNNs). In~\cite{langholz2019oculum} a CNN-based method is proposed for predicting valence and arousal in images by focusing on the ocular region. The reported results in Table~\ref{tab:dimensionalresults} are only RMSE values to have a better comparison with other mentioned works as other methods have only provided this metric in their work. It can  be seen that BReG-NeXt-50 in  overall (considering both valence and arousal) achieves a lower RMSE compared to other methods. 

On AffectNet, the improvement over ResNet and other state-of-the-art methods is significant. Both BReG-NeXt-32 and BReG-NeXt-50 outperform ResNets for arousal and overall predictions. This shows that our complex mapping has been able to fit the training data better and recognize the subtle differences in the dimensional model of affect. For valence, our networks achieve better performance compared to their ResNet counterpart but they do not beat the state-of-the-art. This can be seen in Figure~\ref{fig:errorHist} as well. Where error rates for arousal are concentrated around zero while their corresponding prediction errors for valence are not as dense around zero.

On Affect-in-Wild, our method outperforms ResNet and state-of-the-art methods on all valence, arousal, and overall predictions. This improvement, however, is not as significant as AffectNet. This can be partially due to the fact that the labels for valence and arousal in Affect-in-Wild dataset are very inconsistent in consecutive frames. There are many cases in the dataset that the emotion of the face does not change at all or it changes very subtly, but the labels for the sequence change drastically. Therefore, the network is confused by these type of instances in the training set while on AffectNet -where there is less inconsistency among the labels- our method performs considerably better.

Table~\ref{tab:cc-ccc-sagr} provides additional metrics for the validation set of the studied databases on BReG-NeXt-50. We defined these metrics in Section~\ref{sec:evalmetrics}. On AffectNet, our method achieves high correlation scores for both CC (Equation~\eqref{eq:CC}) and CCC (Equation~\eqref{eq:CCC}) as well as significant sign agreement for both valence and arousal. On Affect-in-Wild, which is a more challenging database in general, the correlation of the  predictions are not high in terms of numbers but are better or comparable with the correlations reported in~\cite{hasani2019bounded} for BReG-Net. For SAGR, however, our method achieves a satisfying prediction which shows that BReG-NeXt is able to correctly predict whether an emotion is either positive or negative as well as whether it is an active emotion or a passive one.

\begin{table}[]
	\centering
		\caption{Results of investigated functions (Equations~\ref{eq:modification_funnctions} and~\ref{eq:arctanmapping}) on BReG-NeXt-50 architecture}
		\label{Tab:resultsoffucntion}
		\resizebox{\columnwidth}{!}{
			\begin{tabular}{cl|c|c|c|c}
				\cline{2-6}
				& \multicolumn{1}{c|}{database} & $\mathcal{H}_1$     & $\mathcal{H}_2$     & $\mathcal{H}_3$     & \begin{tabular}[c]{@{}c@{}}$\mathcal{H}$\\ (Equation~\ref{eq:arctanmapping} )\end{tabular} \\ \hline\hline
				\multicolumn{1}{c|}{\multirow{2}{*}{\begin{tabular}[c]{@{}c@{}}categorical\\ (accuracy)\end{tabular}}} & AffectNet                     & 64.09  & 65.66  & 62.82  & 68.50                                                   \\ \cline{2-6} 
				\multicolumn{1}{c|}{}                                                                                  & FER2013                       & 69.60  & 70.23  & 68.66  & 71.53                                                   \\ \hline\hline
				\multicolumn{1}{c|}{\multirow{2}{*}{\begin{tabular}[c]{@{}c@{}}dimensional\\ (RMSE)\end{tabular}}}     & AffectNet                     & 0.2888 & 0.2833 & 0.2910 & 0.2577                                                  \\ \cline{2-6} 
				\multicolumn{1}{c|}{}                                                                                  & Aff-in-Wild                   & 0.3090 & 0.3008 & 0.3122 & 0.2882                                                  \\ \hline
			\end{tabular}
			}
\end{table}

\begin{figure*}[th]
	\centering
	\subfigure[AffectNet]{\label{fig:AffectNetResutlsVA}\includegraphics[width=.47\linewidth]{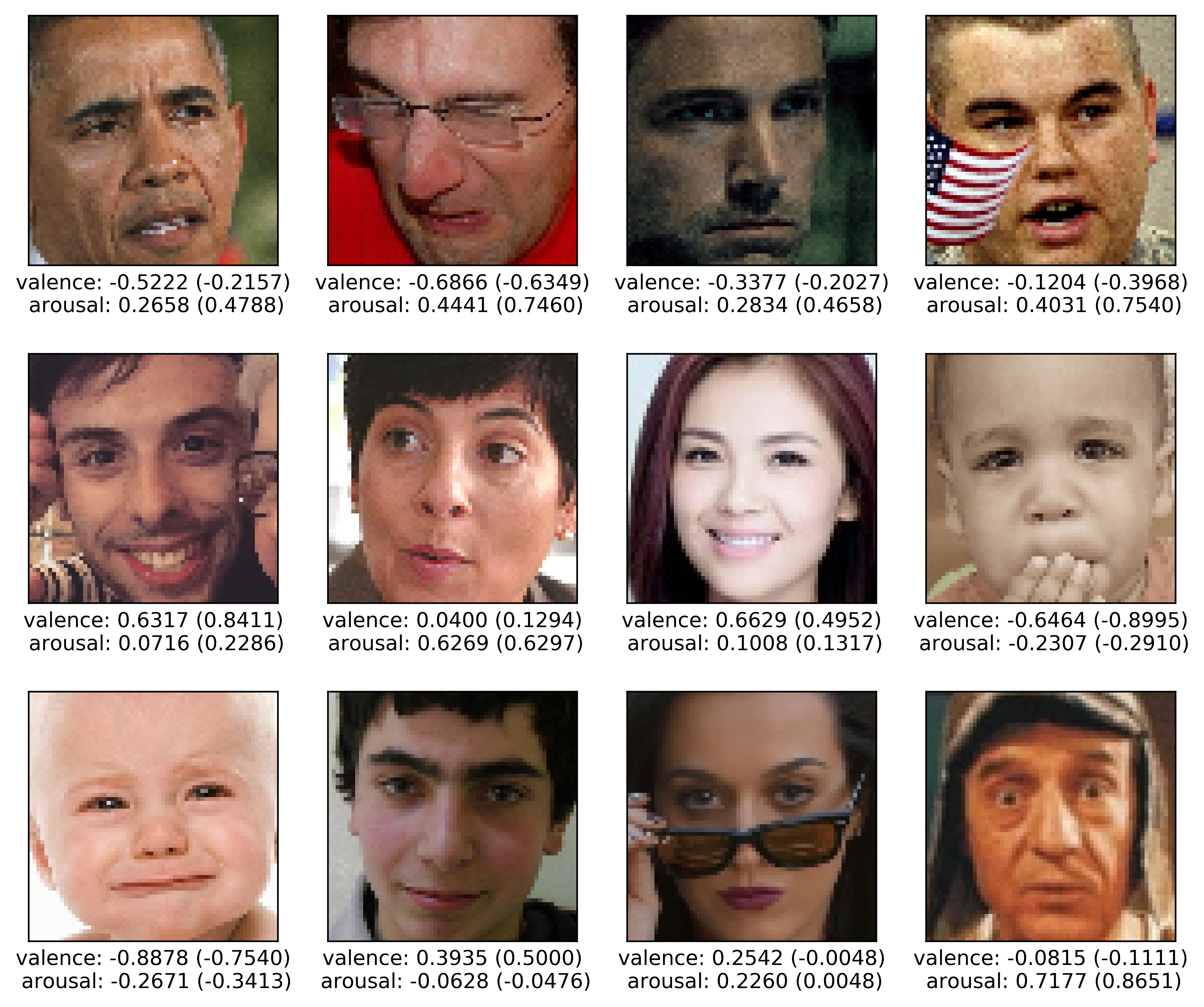}} 
	\qquad
	\subfigure[Affect-in-Wild]{\label{fig:AffinWildResultsVA}\includegraphics[width=.47\linewidth]{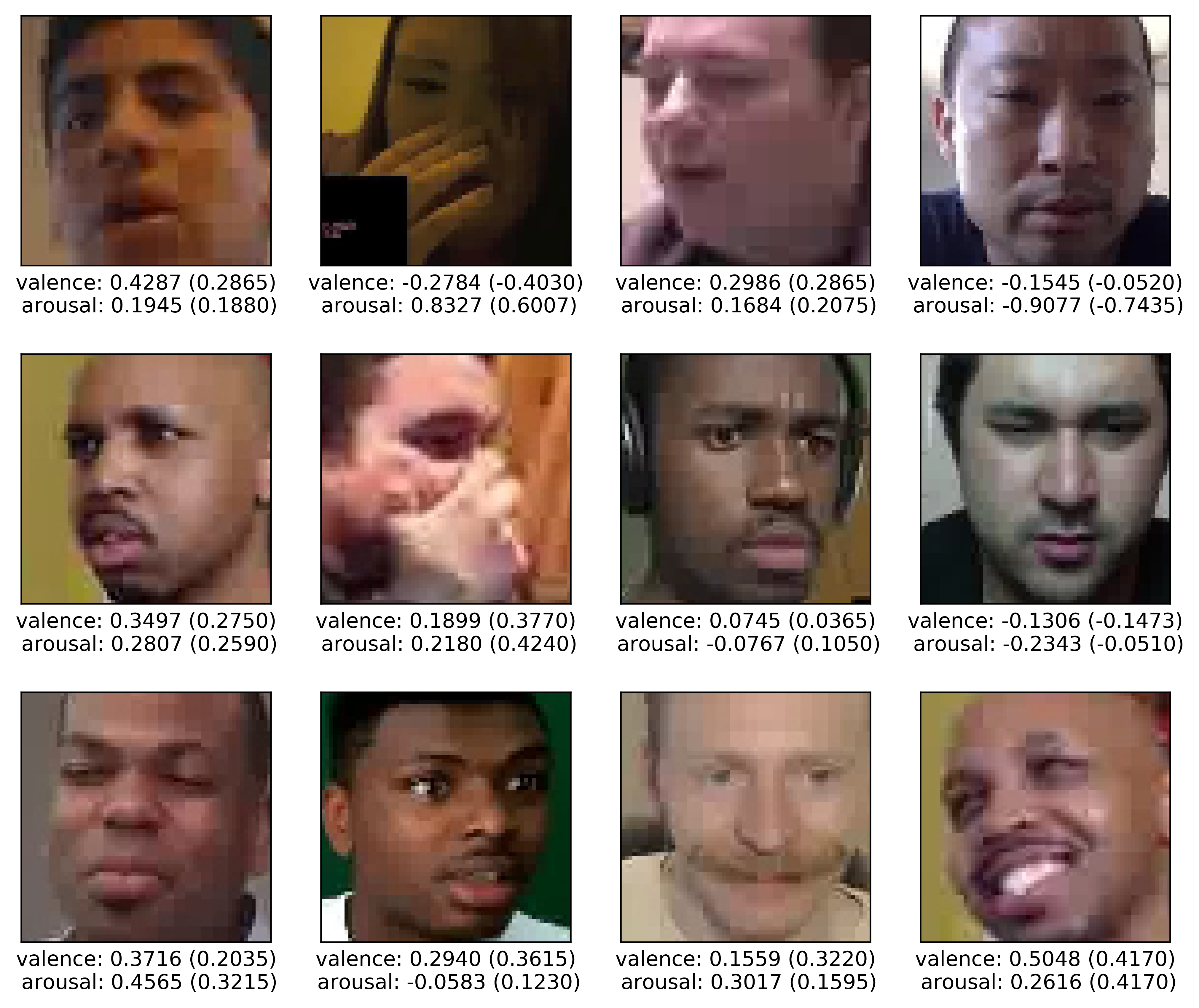}}
	\caption{Example predictions of BReG-NeXt-50 on AffectNet (a) and Affect-in-Wild (b). The text below images indicates predicted values for valence and arousal followed by their corresponding ground-truth in parenthesis.}\label{fig:predictionsVA}
\end{figure*}

Figure~\ref{fig:errorHist} shows the histogram of the prediction errors in the studied databases. In all cases, most error values fall in the vicinity of the zero. On Affect-in-Wild, some high error rates can be seen (especially on arousal predictions) that can be due to the inconsistency of the labels in the training labels as mentioned before. However, in general distribution of the errors have the expected shape and they are mainly gathered around zero on both databases.

Figure~\ref{fig:predictionsVA} shows some examples of BReG-NeXt-50 predictions for the dimensional model of affect. It can be seen that BReG-NeXt is able to recognize the subtle facial muscle shapes for positivity or negativity of expression as predicted values for valence shows in most cases have the same sign (with close value) compared to the ground-truth. Also, BReG-NeXt-50 shows satisfying performance for predicting whether an emotion is active or passive by performing satisfying prediction for arousal. Some of the good examples are surprised or saddened instances in Figure~\ref{fig:predictionsVA}.

\begin{table}[tbp]
	\centering
	
	\caption{Result of non-adaptive networks ($\alpha = 1$, $\beta = 0$)}
	\label{tab:nonadaptiveresults}
	\resizebox{\columnwidth}{!}{
	\begin{tabular}{cl|c|c}
		\cline{2-4}
		\multicolumn{1}{l}{}                                                                                            & \multicolumn{1}{c|}{\textbf{database}} & \textbf{\begin{tabular}[c]{@{}c@{}}BReG-NeXt-32\\ (non-adaptive)\end{tabular}} & \multicolumn{1}{l}{\textbf{\begin{tabular}[c]{@{}l@{}}BReG-NeXt-50\\ (non-adaptive)\end{tabular}}} \\ \hline\hline
		\multicolumn{1}{c|}{\multirow{2}{*}{\textbf{\begin{tabular}[c]{@{}c@{}}categorical\\ (accuracy)\end{tabular}}}} & \textbf{AffectNet}                     & 66.03                                                                          & 67.45                                                                                              \\ \cline{2-4} 
		\multicolumn{1}{c|}{}                                                                                           & \textbf{FER2013}                       & 68.88                                                                          & 70.00                                                                                              \\ \hline\hline
		\multicolumn{1}{c|}{\multirow{2}{*}{\textbf{\begin{tabular}[c]{@{}c@{}}dimensional \\ (RMSE)\end{tabular}}}}    & \textbf{AffectNet}                     & 0.2793                                                                         & 0.2677                                                                                             \\ \cline{2-4} 
		\multicolumn{1}{c|}{}                                                                                           & \textbf{Affect-in-Wild}                & 0.3064                                                                         & 0.2903                                                                                             \\ \hline
	\end{tabular}
}
\end{table}

As mentioned earlier, we investigated several mapping functions and among those Equation~\ref{eq:arctanmapping} showed the best results as this mapping extracts more useful features in each block and is much more flexible in learning the patterns due to the adaptive parameters in each block of the proposed architecture. Table~\ref{Tab:resultsoffucntion} compares the investigated mappings presented in Equations~\ref{eq:modification_funnctions} and~\ref{eq:arctanmapping}. It can be seen that our proposed adaptive complex mapping (Equation~\ref{eq:arctanmapping}) achieves better results in all of the conducted experiments.

In order to show the impact of adaptive complex mapping on facial affect estimation, we evaluated BReG-NeXt with fixed values for $\alpha_l$ and $\beta_l$ in Equation~\eqref{eq:arctanmapping}. Table~\ref{tab:nonadaptiveresults} shows the result of BReG-NeXt-32 and BReG-NeXt-50 when $\forall i \in \mathbb{N} : \alpha_i = 1,~\beta_i = 0$. Therefore, Equation~\eqref{eq:arctanmapping} will be simplified to $\mathcal{H}(\ve{x}_l) = \tan^{-1}(\ve{x}_l)$ in these cases. By comparing the experimental results provided in Tables~\ref{tab:nonadaptiveresults},~\ref{tab:categoricalresults}, and~\ref{tab:dimensionalresults} it can be seen that for all cases adaptive BReG-NeXt outperform their corresponding non-adaptive ones. This improvement is more significant in the dimensional model of affect where subtle changes in the shape of facial muscles result in different values for valence and arousal. This shows that adaptive complex mapping fits each residual unit to its input feature map more than non-adaptive one resulting to extract the subtle meaningful changes in each layer. Our final trained model shows smaller values for $\alpha$ and $\beta$ ($\sim$0.5) in the first few adaptive complex mapping units and larger values ($\sim$1.2) for them in the deeper units in the network. As mentioned before, our code and trained parameters will be publicly available for the research community.

\section{Conclusion}\label{sec:conclusion}

In this work, we introduced BReG-NeXt, a new residual-based network architecture consisting of a  differentiable and bounded gradient function instead of a shortcut path between the input and the output of the residual unit (identity mapping) for the task of affect estimation in both categorical and dimensional models of affect. By utilizing this complex function (called complex mapping), the networks will have more complex nodes and therefore, more useful features are extracted at each layer. Thus, the resulting network is shallower with less number of parameters to learn and fewer operations to perform. 

We showed that our complex mapping needs to have bounded derivative and away from zero for the gradient to flow smoothly in the back-propagation phase which results in preventing from facing vanishing/exploding gradient. It has been shown that incorporating training parameters in the bypass route of residual units results in a better fit especially in challenging tasks such as facial affect estimation where very subtle changes in the training data are needed to be recognized by the network. 

We replaced the identity mapping in original residual units with our adaptive complex mapping in Equation~\eqref{eq:arctanmapping}. Among many other functions that we investigated, this mapping satisfies the required properties for the bypass and it also showed the best results in both affect estimation tasks in our experiments by having a significantly lower number of parameters and FLOPs compared to deep ResNets (Table~\ref{tab:arch}). Furthermore, adding training parameters to the bypass helped to further improve the fitting and be able to distinguish the subtle changes especially in the dimensional model of affect.   

To evaluate our proposed method, we conducted comprehensive experiments for facial affect estimation on categorical and dimensional models of affect. Challenging in-the-wild databases (AffectNet, FER2013, and Affect-in-Wild) were used for the experiments. We showed that our adaptive complex mapping outperforms original residual units with identity mapping and other state-of-the-art methods in the field in the majority of the cases (Tables~\ref{tab:categoricalresults} and~\ref{tab:dimensionalresults}). Considering the trade-off between the number of parameters and recognition rate, we proposed BReG-NeXt-50 as our final architecture for this task. Furthermore, we provided additional metrics in both affect models to have a better evaluation of our method. In the categorical model, BReG-NeXt-50 with only 3.1M training parameters, achieves 68.50\% and 71.53\% accuracy on AffectNet and FER2013 databases, respectively. And in the dimensional model, it achieves 0.2577 and 0.2882 for RMSE on AffectNet and Affect-in-Wild databases, respectively.

A recommendation for future work is to apply this method to other residual-based networks in DNNs. For instance, BReG-NeXt architecture can be applied to the well-known DenseNet~\cite{huang2017densely} architecture to enrich the feature map flowing through the network. However, memory and computational power limitations need to be considered in such networks with nested connections. Also, for a more comprehensive investigation, BReG-NeXt architecture can be expanded furthermore with 3D-CNNs with approaches similar to~\cite{vielzeuf2017temporal}.

\ifCLASSOPTIONcompsoc
  \section*{Acknowledgments}
\else
  \section*{Acknowledgment}
\fi
This work is partially supported by the NSF grant CNS-1427872. We gratefully acknowledge the support of NVIDIA Corporation with the donation of the TITAN Xp GPU used for this research.

\ifCLASSOPTIONcaptionsoff
  \newpage
\fi

\bibliography{egbib}

\vspace{-0.4cm}
\begin{IEEEbiography}[{\includegraphics[width=1in,height=1.25in,clip,keepaspectratio]{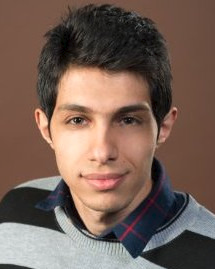}}]{Behzad Hasani}
received the BSc degree in computer hardware engineering from Khaje Nasir Toosi University of Technology, Tehran, Iran, in 2013, and the MSc degree in computer engineering - artificial intelligence from Iran University of Science and Technology, Tehran, Iran, in 2015. He is currently pursuing his Ph.D. degree in electrical \& computer engineering and is a graduate research assistant in the Department of Electrical and Computer Engineering at the University of Denver. His research interests include Computer Vision, Machine Learning, and Deep Neural Networks, especially on facial expression analysis.
\end{IEEEbiography}

\begin{IEEEbiography}[{\includegraphics[width=1in,height=1.25in,clip,keepaspectratio]{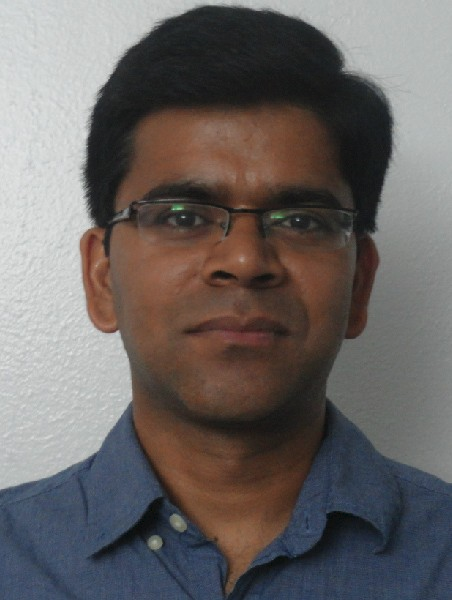}}]{Pooran Singh Negi} received the BSc degree in computer engineering from GBPUAT, India in 2000, and the Ph.D. degree in mathematics in 2012 from University of Houston. He is currently a visiting assistant professor of Data Science in the computer science department at the University of Denver. His research interests include Applied Harmonic Analysis, Information Theory, Machine Learning, and Deep Neural Networks.

\end{IEEEbiography}

\begin{IEEEbiography}[{\includegraphics[width=1in,height=1.25in,clip,keepaspectratio]{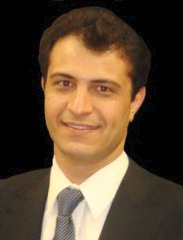}}]{Mohammad H. Mahoor}
	
	received the MS degree in biomedical engineering from Sharif University of Technology, Iran, in 1998, and the Ph.D. degree in electrical and computer engineering from the University of Miami, Florida, in 2007. Currently, he is a proffesor of Electrical and Computer Engineering at the University of Denver. He does research in the area of computer vision and machine learning including visual object recognition, object tracking, affective computing, and human-robot interaction (HRI) such as humanoid social robots for interaction and intervention with children with autism and older adults with depression and dementia. He has received over \$7M of research funding from state and federal agencies including the National Science Foundation and the National Institute of Health. He is a Senior Member of IEEE and has published over 125 conference and journal papers.
	
\end{IEEEbiography}

\end{document}